\documentclass[lettersize,journal]{IEEEtran}
\usepackage{amsmath,amsfonts}
\usepackage{algorithmic}
\usepackage{algorithm}
\usepackage{array}
\usepackage[caption=false,font=normalsize,labelfont=sf,textfont=sf]{subfig}
\usepackage{textcomp}
\usepackage{stfloats}
\usepackage{url}
\usepackage{verbatim}
\usepackage{graphicx}
\usepackage{cleveref}
\usepackage{cite}
\usepackage{multirow}

\usepackage[small]{caption}
\usepackage{booktabs}
\usepackage{makecell}
\usepackage{comment}
\usepackage{amssymb} % define this before the line numbering.
\usepackage{subcaption}
\usepackage{multirow}
\usepackage{wrapfig}
\usepackage{color} 
\usepackage{xcolor} 
\usepackage{enumerate}
\usepackage{pifont}

\begin{document}

% \title{Efficient Active Training for \textcolor{red}{Generalized} Deep LiDAR Odometry}
\title{Efficient Active Training for Deep LiDAR Odometry}

\author{Beibei Zhou,  Zhiyuan Zhang, Zhenbo Song, Jianhui Guo, Hui Kong*
        % <-this % stops a space
\thanks{Beibei Zhou is with Shanghai Polytechnic University, Shanghai, China (e-mail: beibeizhou18@gmail.com).
Zhiyuan Zhang is with Singapore Management University, Singapore (e-mail: cszyzhang@gmail.com).
Zhenbo Song and Jianhui Guo are with Nanjing University of Science and Technology, Nanjing, Jiangsu (\{songzb,guojianhui\}@njust.edu.cn).
Hui Kong is with University of Macau, Macau, China (e-mail: huikong@um.edu.mo). 
}
\thanks{*Correspondence author.}}

\markboth{Journal of \LaTeX\ Class Files,~Vol.~14, No.~8, August~2021}%
{Shell \MakeLowercase{\textit{et al.}}: A Sample Article Using IEEEtran.cls for IEEE Journals}


\maketitle

\begin{abstract}

Robust and efficient deep LiDAR odometry models are crucial for accurate localization and 3D reconstruction, but typically require extensive and diverse training data to adapt to diverse environments, leading to inefficiencies. To tackle this, we introduce an active training framework designed to selectively extract training data from diverse environments, thereby reducing the training load and enhancing model generalization.
Our framework is based on two key strategies: Initial Training Set Selection (ITSS) and Active Incremental Selection (AIS). ITSS begins by breaking down motion sequences from general weather into nodes and edges for detailed trajectory analysis, prioritizing diverse sequences to form a rich initial training dataset for training the base model. 
For complex sequences that are difficult to analyze, especially under challenging snowy weather conditions, AIS uses scene reconstruction and prediction inconsistency to iteratively select training samples, 
refining the model to handle a wide range of real-world scenarios.
Experiments across datasets and weather conditions validate our approach's effectiveness. Notably, our method matches the performance of full-dataset training with just 52\% of the sequence volume, demonstrating the training efficiency and robustness of our active training paradigm. By optimizing the training process, our approach sets the stage for more agile and reliable LiDAR odometry systems, capable of navigating diverse environmental conditions with greater precision.
\end{abstract}


\section{Introduction}

\IEEEPARstart{L}iDAR odometry (LO) is a foundational technique for Simultaneous Localization and Mapping (SLAM)~\cite{Matsuki_2024_CVPR,10681669,10938051}, which is widely used in the field of autonomous driving. It underpins critical applications demanding robust environmental perception and real-time localization, including navigation, 3D mapping, and augmented reality. However, LO performance degrades severely under adverse weather conditions, particularly snowfall, critically undermining system reliability. While prevailing strategies attempt to bolster adaptability by training models on extensive, multi-condition datasets, these approaches encounter inherent limitations: prohibitive computational costs, protracted training durations, and inefficiencies arising from non-discriminative data sampling across these diverse datasets. Therefore, developing a generalized and computationally efficient training framework is paramount to ensuring robust LO performance in complex environments.

To ensure an effective and robust LO network, it is essential to minimize redundant diverse data and maintain a balanced representation of motion patterns. Although extreme weather conditions are rare, their inclusion in the training set is crucial due to their significant impact on the performance of autonomous systems. These corner cases are vital for developing an LO network capable of handling real-world driving scenarios with reliability and safety.
Current research on efficient training and generalization of deep autonomy models primarily focuses on perception~\cite{zhang2023perception,carballo2020libre,mohammed2020perception}, detection~\cite{gupta2024robust,jia2022multi,jia2023hdgt,shi2022motion,vora2020pointpainting}, and planning~\cite{lu2024activead}, with comparatively less attention given to localization and motion estimation. In the field of deep LO, GRAMME~\cite{almalioglu2022deep} enhances model generalization under adverse conditions by integrating multi-sensor data. However, this approach increases hardware costs and leads to larger datasets and longer training times, further complicating the challenge of efficient training.

\begin{figure*} [t] 
\centering  
\includegraphics[width=0.93\linewidth]{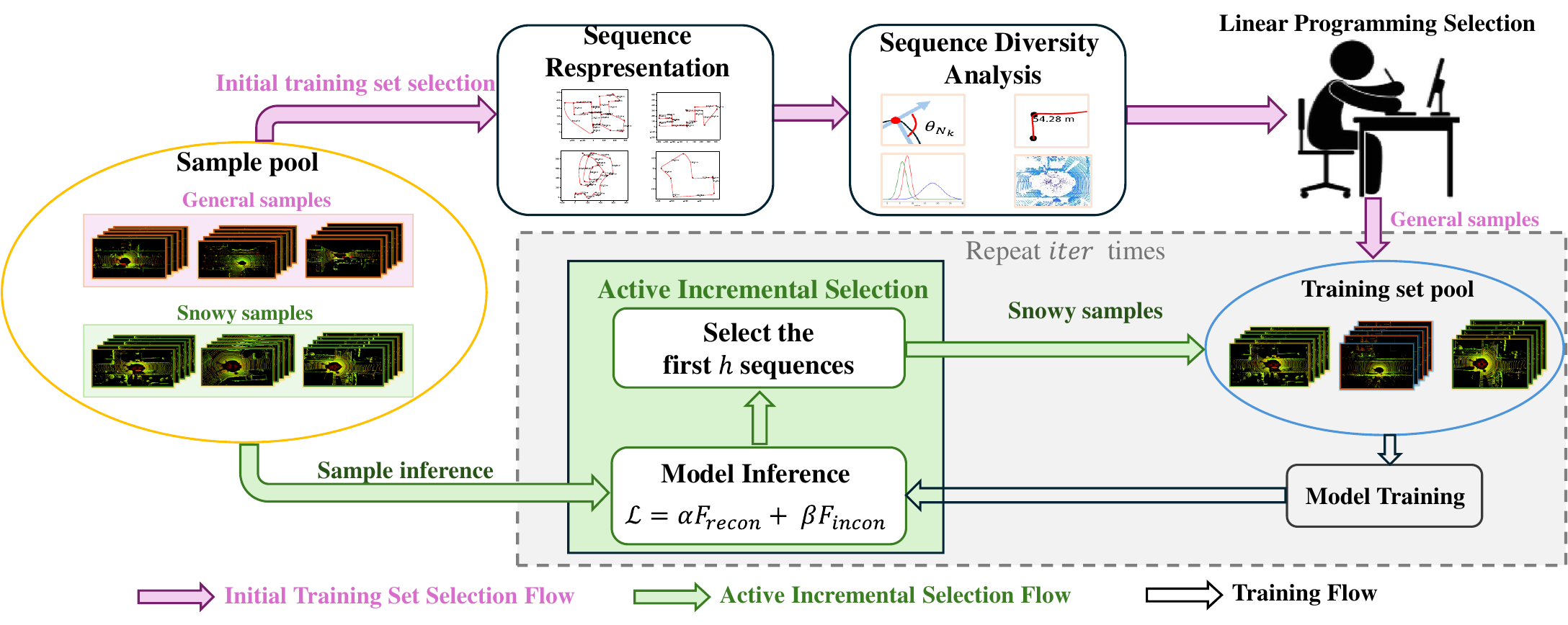}  
		\caption{The overview of ActiveLO-training process. We analyze the sequence diversity of general samples with complex motion states and add the most informative sequences into the initial training set pool. Then, the model employs the incremental selection strategy to actively identify and select hard snow sequences, and this process is iterated for $iter$ times until the model's performance is optimized.}  
\label{fig:overview}  
\vspace{-0.3cm}
\end{figure*}

This paper introduces ActiveLO-training, a novel active sample selection strategy designed to enhance the training efficiency and robustness of the deep learning-based LiDAR Odometry (LO) models. Our objective is to achieve accuracy comparable to models trained on full datasets while significantly reducing the required training sample size. ActiveLO-training comprises two core components: Initial Training Set Selection (ITSS) and Active Incremental Selection (AIS). ITSS constructs a compact, foundational training set by capturing a broad spectrum of motion states under general weather conditions, establishing foundational pose estimation capabilities for the base model. AIS then strategically integrates challenging samples, particularly those from adverse conditions like extreme snowy weather, to enhance robustness. To our knowledge, this represents the first active training set selection framework specifically developed for deep LO applications.

The ITSS module employs a targeted approach to maximize motion pattern diversity. It segments training trajectories into nodes and edges based on curvature, enabling granular analysis of dynamics through features like velocity changes, turning angles, and so on. This allows ITSS to quantify trajectory variability and strategically select an initial, compact set rich in diverse motion states. However, ensuring robustness against complex environmental conditions—especially adverse snowy weather that introduces significant noise into LiDAR point clouds—presents a distinct challenge. Such scenarios are inherently difficult to quantify, often underrepresented in datasets, and critically impact model generalizability. To address this, AIS actively identifies and incorporates these challenging, underrepresented samples essential for resilience. It quantifies sample importance based on scene reconstruction loss and prediction inconsistency loss.
By progressively enriching the dataset with these critical examples, AIS significantly improves the LO network's accuracy and robustness across diverse environmental challenges.

To summarize, our contributions mainly include:
\begin{itemize}

    \item 
    \textbf{Diverse Motion Pattern Selection:} We propose an initial training set selection strategy that ensures a comprehensive representation of 6-Degree-of-Freedom (6-DoF) motion patterns, based on a novel quantitative evaluation of trajectory variability and significance.
    
    \item
    \textbf{Iterative Hard Sample Incorporation:} We present an incremental selection strategy that enriches the training set with challenging samples, identified through unsupervised metrics such as scene reconstruction loss and prediction inconsistency loss, thereby enhancing the robustness of the model.
    
    \item 
    \textbf{Performance Benchmarking:} Our extensive experimental evaluation demonstrates that ActiveLO training not only surpasses random sample selection but also achieves performance parity with the utilization of the full dataset, with the remarkable efficiency of employing only 52\% of the training sequences.
    
\end{itemize}

\vspace{-0.3cm}
\section{Related Work}
\label{sec:Related Work}

\subsection{Supervised LiDAR Odometry}

In the realm of supervised LiDAR odometry, a multitude of methodologies have been proposed to project 3D point clouds onto 2D planes for enhanced computational efficiency and accuracy. Notably, Velas et al.~\cite{velas2018cnn} conceptualized odometry as a classification task, yielding dense matrix outputs for precise localization. DeepPCO~\cite{wang2019deeppco} pioneered a dual-network architecture aimed at refining both translation and orientation estimation with high precision. 
Subsequent approaches, such as DMLO~\cite{li2020dmlo}, LodoNet~\cite{zheng2020lodonet}, and the work of Liu et al.~\cite{liu2022lidar}, have employed 2D projections coupled with additional layers to bolster keypoint matching, thereby increasing the complexity and computational demands of the networks. 
In contrast, LO-Net~\cite{li2019net}, PWCLO-Net~\cite{wang2021pwclo}, and DELO~\cite{ali2023delo} have opted to directly manipulate 3D point clouds, bypassing the need for 2D projections. LO-Net~\cite{li2019net}, in particular, integrates normal estimation and mask prediction modules within its architecture, utilizing a cylindrical projection to enhance feature extraction. Furthermore, PWCLO-Net~\cite{wang2021pwclo} and EfficientLO-Net~\cite{wang2022efficient} have adopted multiscale, layered network structures that effectively obviate the need for keypoint matching, which is traditionally considered a cornerstone of precision in supervised LO tasks. DELO~\cite{ali2023delo} integrates partial optimal transport into the LO framework, enhancing frame-to-frame correspondence accuracy and incorporating uncertainty for reliable predictions.

\vspace{-0.3cm}
\subsection{Unsupervised LiDAR Odometry}

The advent of unsupervised methods in LiDAR odometry~\cite{cho2020unsupervised, 10906337,cho2019deeplo, SelfVoxeLO, xu2022robust} has garnered significant attention, primarily due to their ability to mitigate the exigency of labor-intensive manual annotations. Cho et al.~\cite{cho2020unsupervised}, expanding upon their seminal work~\cite{cho2019deeplo}, introduced one of the pioneering unsupervised deep Learning-based Odometry systems. This system is underpinned by a geometric perception consistency loss, drawing inspiration from the point-to-plane Iterative Closest Point (ICP) algorithm~\cite{low2004Linear}. 
Subsequently, Nubert et al.~\cite{nubert2021self} enhanced this framework by incorporating plane-to-plane losses, thereby refining the alignment of points with their corresponding normal vectors. SelfVoxeLO~\cite{SelfVoxeLO} employs a 3D convolutional network directly on raw LiDAR point clouds, extracting salient features and leveraging multiple self-supervised losses for robust feature learning.
RSLO~\cite{xu2022robust} presents a two-stage odometry network architecture, which initially estimates self-motion within subregions using 3D CNNs, followed by a sophisticated voting mechanism for pose selection and structural representation. Li et al.~\cite{li20233d} proposed a coarse-to-fine strategy to facilitate scene flow estimation for pose correction; however, this approach does not entirely resolve the issue of establishing strict correspondences. SSLO~\cite{fu2022self} employs a variety of loss functions to address matching challenges in 2D projections, yet it encounters limitations due to suboptimal feature utilization.
In a significant advancement, HPPLO-Net~\cite{zhou2023hpplo} integrates scene flow estimation, differentiable weighted Point-to-Plane Singular Value Decomposition (SVD), and inlier mask prediction. This integrated approach effectively mitigates the challenges associated with data association in sparse environments, thereby enhancing the robustness of unsupervised LiDAR odometry.

\vspace{-0.3cm}
\section{Approach}
Our approach is a two-stage training set selection strategy, comprising an initial selection phase and an incremental enhancement phase. The initial phase involves extracting a baseline training set from the KITTI Odometry dataset under typical weather conditions. The subsequent phase integrates snow-affected sequences from the CADC dataset into the existing training set, incrementally expanding the model's exposure to diverse conditions. The details of these stages are delineated below.

\vspace{-0.3cm}
\subsection{Initial Training Set Selection Strategy}
\subsubsection{Sequence Representation}

For the purpose of selecting a diverse set of training sequences, we segment the trajectory of each sequence into discrete nodes and edges. This segmentation allows us to compute their respective features utilizing the dataset-provided information, thereby facilitating the selection of representative training samples.

\begin{figure} [t] 
\centering  
\includegraphics[width=0.9\linewidth]{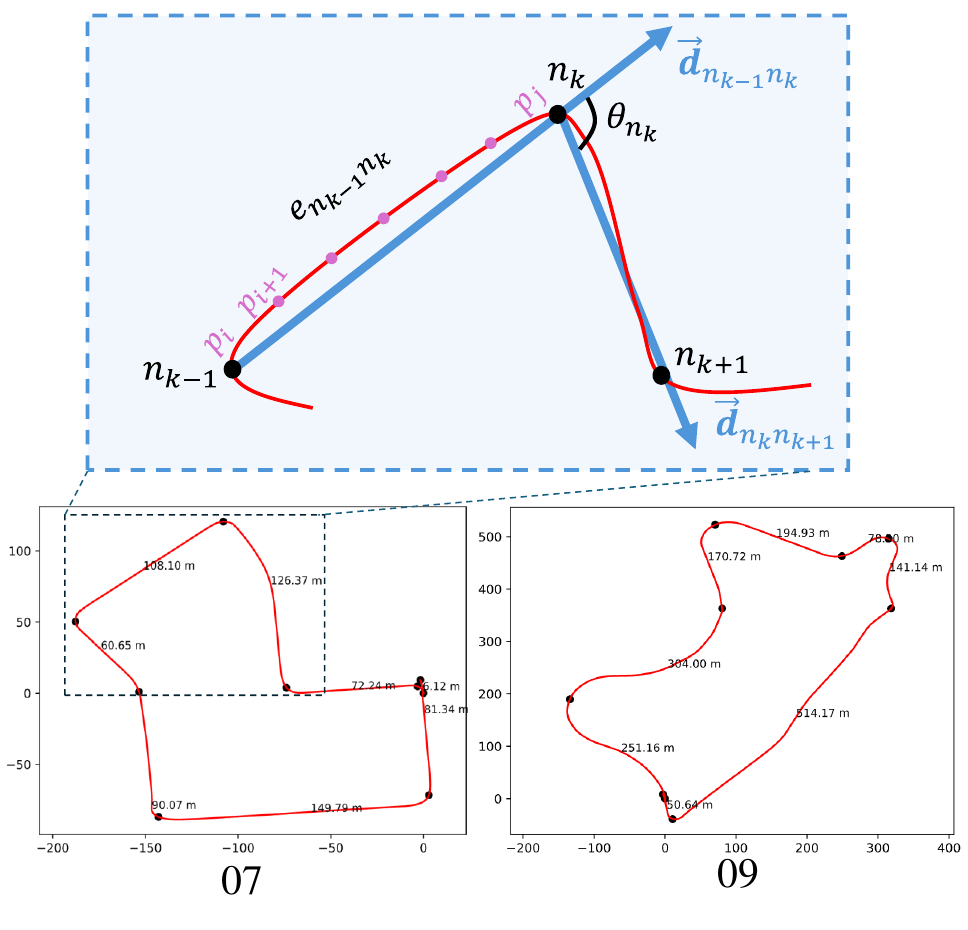}  
		\caption{Labeling results for nodes and edges of sequences 07 and 09. Black dots indicate labeled nodes, and the length of edges between nodes is expressed in meters (m). The detailed calculation process of node features is shown in the blue box.}  
\label{traj_cal}  
\vspace{-0.3cm}
\end{figure}

\textit{a)~Trajectory Nodes and Edges.}
Let $\mathbf{P}_i = \{\mathbf{p}_i^{j} = (x_i^{j}, y_i^{j}, z_i^{j}) \mid j=0,1,\dots,n\}$ denote the point cloud of frame $i$ in a sequence, where $x_i^{j}$, $y_i^{j}$, and $z_i^{j}$ are the Cartesian coordinates. We set the origin at the first frame's coordinates and the endpoint at the last frame of the sequence. A frame is designated as a node from a selection of neighboring frames whenever the vehicle navigates a turn. The set of nodes is denoted as $\mathbf{N}=\{\mathbf{n}_{k} \mid k=0,1,\dots, m\}$, where $\mathbf{n}_{k}$ represents the coordinates of node $k$. 
An edge between two nodes, denoted as $\mathbf{e}_{\mathbf{n}_{k-1}\mathbf{n}_{k}} = \{\mathbf{p}_{f} \mid f=i,i+1,\dots, j\}$, is defined as the sequence of coordinates from frame $i$ to frame $j$. This results in exactly $j-i+1$ frames bridging the two nodes. Each node, excluding the starting and ending points, has an in-degree and an out-degree of 1. Examples of labeled results are shown in \cref{traj_cal}.

\textit{b)~Node Features.} The node feature is defined by the vehicle's turning angle at that node.
The angle is calculated based on the vectors formed between the node and its two adjacent nodes. As illustrated in~\cref{traj_cal}, the nodes adjacent to node $\textbf{n}_k$ are $\textbf{n}_{k-1}$ and $\textbf{n}_{k+1}$. The distance vectors between these nodes are defined as  $\vec{\mathbf{d}}_{\textbf{n}_{k-1}\textbf{n}_{k}} = \textbf{n}_{k} - \textbf{n}_{ k-1} $ and $\vec{\mathbf{d}}_{\textbf{n}_{k}\textbf{n}_{k+1}} = \textbf{n}_{k+1} - \textbf{n}_{k} $. Subsequently, the angle between two adjacent vectors $\theta_{\textbf{n}_{k}}$ at node $\textbf{n}_k$ is computed as
% \small
\begin{equation}
% \begin{aligned}
\theta_{\textbf{n}_{k}} = \arccos(\frac{\vec{\mathbf{d}}_{\textbf{n}_{k-1}\textbf{n}_{k}} \cdot \vec{\mathbf{d}}_{\textbf{n}_{k}\textbf{n}_{k+1}}}{\|\vec{\mathbf{d}}_{\textbf{n}_{k-1}\textbf{n}_{k}}\|_2 \|\vec{\mathbf{d}}_{\textbf{n}_{k}\textbf{n}_{k+1}}\|_2}) \cdot \frac{\pi}{180}.\\ \label{node_feature}
% \end{aligned}
\end{equation}

Note that the first and last nodes lack angles, so we can calculate the mean and standard deviation as follows,
% \small
\begin{equation}
\small
% \begin{aligned}
\overline{\theta}= \frac{1}{m-1} \sum_{i=1}^{m-1} \theta_{\textbf{n}_k},
\sigma_\theta= \sqrt{\frac{1}{m-1} \sum_{i=1}^{m-1} (\theta_{\textbf{n}_k} - \overline{\theta})^2}. \label{radian_mean}
% \end{aligned}
\end{equation}
The larger the standard deviation of node radians, the more variable the vehicle's turning angles. In the odometry task, where the model predicts the 6-DoF relative pose between frames, this diversity enhances the model's adaptability to various rotations.

\textit{c)~Edge Features.} 
Assuming that the vehicle's driving conditions (such as speed and direction) are consistent within each edge, we can characterize these edges by the trajectory length and average speed. The length $l_{\textbf{n}_{k}\textbf{n}_{k+1}}$ and the speed $v_{\textbf{n}_{k}\textbf{n}_{k+1}}$ of edge $\mathbf{e}_{\textbf{n}_{k}\textbf{n}_{k+1}}$ is calculated as follows:
% \small
\begin{equation}
\small
l_{\textbf{n}_{k}\textbf{n}_{k+1}} = \sum_{f=i}^{j-1} \|{p}_f - {p}_{f+1}\|_2, 
v_{\textbf{n}_{k}\textbf{n}_{k+1}} = \frac{l_{\textbf{n}_{k}\textbf{n}_{k+1}}}{(j-i+1) / r},
\end{equation}
where $r$ denotes the LiDAR's frame frequency in Hz.

We define the edges between every two nodes as straight lines, where the edge length represents the travel distance under a consistent motion state. The speed of an edge indicates the diversity of the translation part of 6-DoF poses. Higher speeds result in larger translations and fewer inter-frame point cloud overlaps, while slower speeds lead to smaller translations and more overlaps. To measure sequence diversity, we calculated the mean speed $\overline{v}$ and standard deviation $\sigma_v$ of all edges, as shown in (\cref{speed_mean_var}):
% \small
\begin{equation}
\small
% \begin{aligned}
\overline{v} = \frac{1}{m} \sum_{i=0}^{m-1} v_{\textbf{n}_i\textbf{n}_{i+1}},\\
\sigma_v = \sqrt{\frac{1}{m} \sum_{i=0}^{m-1} (v_{\textbf{n}_i\textbf{n}_{i+1}} - \overline{v})^2}. \label{speed_mean_var}
% \end{aligned}
\end{equation}
And the mean length $\overline{l}$ and standard deviation $\sigma_l$ are calculated as:
% \small
\begin{equation}
\small
% \begin{aligned}
\overline{l} = \frac{1}{m} \sum_{i=0}^{m-1} l_{\textbf{n}_i\textbf{n}_{i+1}},\\
\sigma_l = \sqrt{\frac{1}{m} \sum_{i=0}^{m-1} (l_{\textbf{n}_i\textbf{n}_{i+1}} - \overline{l})^2}. \label{length_mean_var}
% \end{aligned}
\end{equation}

\textit{d)~Sample Features.} 
Besides the vehicle's motion state, we also consider scene complexity by quantifying the proportion of outliers.
Assume that two successive point clouds in the sequence are represented as $\mathbf{Q}_i$ and $\mathbf{Q}_{i+1}$, and the relative pose of them is $\mathbf{T}_{i -> i+1}=[\mathbf{R},\mathbf{t}]$, where $\mathbf{R},\mathbf{t} $ denote the rotation and translation matrices, respectively.
Then we transform $\mathbf{Q}_i$ into $\mathbf{Q}'_{i+1} = \mathbf{R}\mathbf{Q}_i + \mathbf{t}$.
The set of outliers in $\mathbf{Q}_i$ is defined as
% \small
\begin{equation}
\small
% \begin{aligned}
\mathbf{Q}^o_i = \{ \mathbf{q}_i^k \mid \|\mathbf{q}_i^k - NN(\mathbf{q}_{i+1}^{'k})\|_2 > \epsilon \},\\
% \end{aligned}
\end{equation}
where $\mathbf{q}_i^k$ denotes the $k$th point in $\mathbf{Q}_i$,
$\mathbf{q}_{i+1}^{'k}$ denotes the corresponding point in $\mathbf{Q}'_{i+1}$.
$NN(\mathbf{q}_{i+1}^{'k})$ denotes the nearest neighbor point of $\mathbf{q}_{i+1}^{'k}$ in $\mathbf{Q}'_{i+1}$, which is considered as an outlier when the distance is larger than a threshold $\epsilon$.
Subsequently, the outlier proportion of $\mathbf{Q}_i$ is defined as $o_i$, and the total outlier proportion of a sequence is defined as $s_o$:
% \small
\begin{equation}
\small
\begin{aligned}
o_i = \frac{Num(\mathbf{Q}^o_i)}{Num(\mathbf{Q}_i)}, 
s_o = \frac{\sum {Num(\mathbf{Q}^o_i)}}{\sum Num(\mathbf{Q}_i)}, 
\label{inlier_scale}
\end{aligned}
\end{equation}
where $Num(\mathbf{Q}^o_i)$ and $Num(\mathbf{Q}_i)$ represent the number of outliers and the number of all points in the $i$th frame, respectively.

\subsubsection{Sequence Diversity Analysis}
% \label{seq_variegation}
Based on the features calculated above, we develop two metrics to ensure sequence diversity: trajectory variability and trajectory importance.

\textit{a)~Trajectory Variability}. Trajectory variability refers to how much a vehicle's motion state changes over a given sequence. 
Sequences with higher variability can enhance the model's generalization and better adapt to diverse real-world environments.
Consequently, we use the standard deviation of node radian $\sigma_\theta$, edge lengths $\sigma_l$, and edge speed $\sigma_v$ together to describe the trajectory variability of the sequence $s$ as shown in (\cref{var}):
\begin{equation}
% \begin{aligned}
F_{Var}(s) = \lambda_1\sigma_\theta+\lambda_2\sigma_l+\lambda_3\sigma_v,
\label{var}
% \end{aligned}
\small
\end{equation}
where $\lambda_1$, $\lambda_2$, and $\lambda_3$ are the weighting factors.
The larger the variability metrics, the more likely it is selected as the initial training set.

\textit{b)~Trajectory Importance.} 
To ensure the model comprehensively learns various rotation angles and translation distances, the selected training sets must include adequate turns and translations. Thus the importance of a sequence is denoted as the weighted sum of node angles and edge lengths. A higher sum indicates greater importance. Mathematically, the importance $F_{Impor}(s)$ of the sequence $s$ is computed as follows:
% \small
\begin{equation}
\small
\begin{aligned}
F_{Impor}(s) = \lambda_4 \sum_{k=1}^{m-1} \| v_{\textbf{n}_{k-1}\textbf{n}_k} - v_{\textbf{n}_{k}\textbf{n}_{k+1}} \| \theta_{\textbf{n}_k} + \\
\lambda_5 \frac{\sum_{k=1}^{m-1} l_{\textbf{n}_{k}\textbf{n}_{k+1}}}{\sum_{s \in \mathbf{S}} \sum_{k=1}^{m-1} l_{\textbf{n}_{k}\textbf{n}_{k+1}}}. \label{importance}
\end{aligned}
\end{equation}
Here, the first term represents the weighted sum of all angles of the nodes $\theta_{\textbf{n}_i}$ in the sequence $s$, where the weight $v_{\textbf{n}_{i-1}\textbf{n}_i}-v_{\textbf{n}_{i}\textbf{n}_{i+1}}$ corresponds to the speed variation at each node. 
The latter term reflects the proportion of the length of the trajectory relative to the total length of the dataset; a larger proportion indicates the trajectory of greater importance.
$\lambda_4$ and $\lambda_5$ are the weighting factors.

\subsubsection{Linear Programming Selection Process}
% \label{linear}
In the selection process, sequences are first arranged in ascending order of the outliers proportion $s_o$ and average speed $\Bar{v}$, respectively. The training sequences are then divided into several intervals based on the values of $s_o$ and $\Bar{v}$. The requirement that at least one sample sequence must be selected from each interval serves as a constraint in the linear programming model.
By maximizing the values of sequence diversity, \textit{that is,} the weighted addition of trajectory variability and importance, we ultimately select $u$ sequences from the training set pool by the Objective Function (\cref{function}):
\begin{equation}
    Maximize \ \mathbf{Z}=\sum^{10}_{s=0}(F_{Var}(s) x_s + F_{Impor}(s) x_s).
    \label{function}
    \small
\end{equation}

\vspace{-0.3cm}
\subsection{Incremental Selection Strategy}

In order to generalize the model to different levels of snowfall weather, we gradually add snowy samples to the training set.
Since the snow sequences have complex scenarios, and they are difficult to quantify like typical sequences,  
we develop an incremental selection strategy. This strategy enables the model to actively select effective training sequences based on its current predictive ability, guided by two metrics: scene reconstruction loss and prediction inconsistency loss. 
As shown in~\cref{fig:overview}, at the end of each iteration, the intermediate model performs inference on the remaining sequences to compute the two specified metrics for each sequence and subsequently make a selection based on the inference results.
The higher the metric value, the poorer the model's predictive performance on these scenarios, which should therefore be prioritized to be added to the training set pool.

\subsubsection{Scene Reconstruction Loss}
We use the point-to-plane loss function for scene reconstruction within a pair of LiDAR frames $\{ \mathbf{Q_i}, \mathbf{Q_{i+1}} \}$ in a given sequence:
% \small
\begin{equation}
\small
\begin{aligned}
F_{Recon}(\mathbf{Q_i}, \mathbf{Q_{i+1}}) = \sum_{\mathbf{q}_{i}\in \mathbf{Q_i}} \Big[
& f_{(\Phi, \mathbf{R})}(\mathbf{Q_i}, \mathbf{Q_{i+1}}) \mathbf{q}_{i} + \\ 
f_{(\Phi, \mathbf{t})}(\mathbf{Q_i}, \mathbf{Q_{i+1}}) 
& - \mathbf{q}_{i+1} \Big] \cdot \mathbf{n}_{\mathbf{q}_{i+1}}, 
\label{F_loss}
\end{aligned}
\end{equation}
where $f_{(\Phi, \mathbf{R})}$ and $f_{(\Phi, \mathbf{t})}$ denote the model's predicted rotation and translation matrices of the sample pair $\{ \mathbf{Q_i}, \mathbf{Q_{i+1}} \}$, respectively, $\Phi$ denotes the model parameters, 
$\mathbf{q}_{i+1}\in{\mathbf{Q_{i+1}}}$ denotes the nearest neighbor found on $\mathbf{Q_{i+1}}$ of $\mathbf{q}_{i}$, and $\mathbf{n}_{q_{i+1}}$ denotes the normal vector of $\mathbf{q}_{i+1}$.
This loss measures the network's ability to predict the relative pose of the original sample pairs by quantifying the discrepancy between the transformed source frame and the target frame, projected along their local normal direction.
 A higher loss indicates poor alignment under the predicted pose, signaling the network to focus on these challenging samples during training.

\begin{figure} [t] 
\centering  
\includegraphics[width=0.98\linewidth]{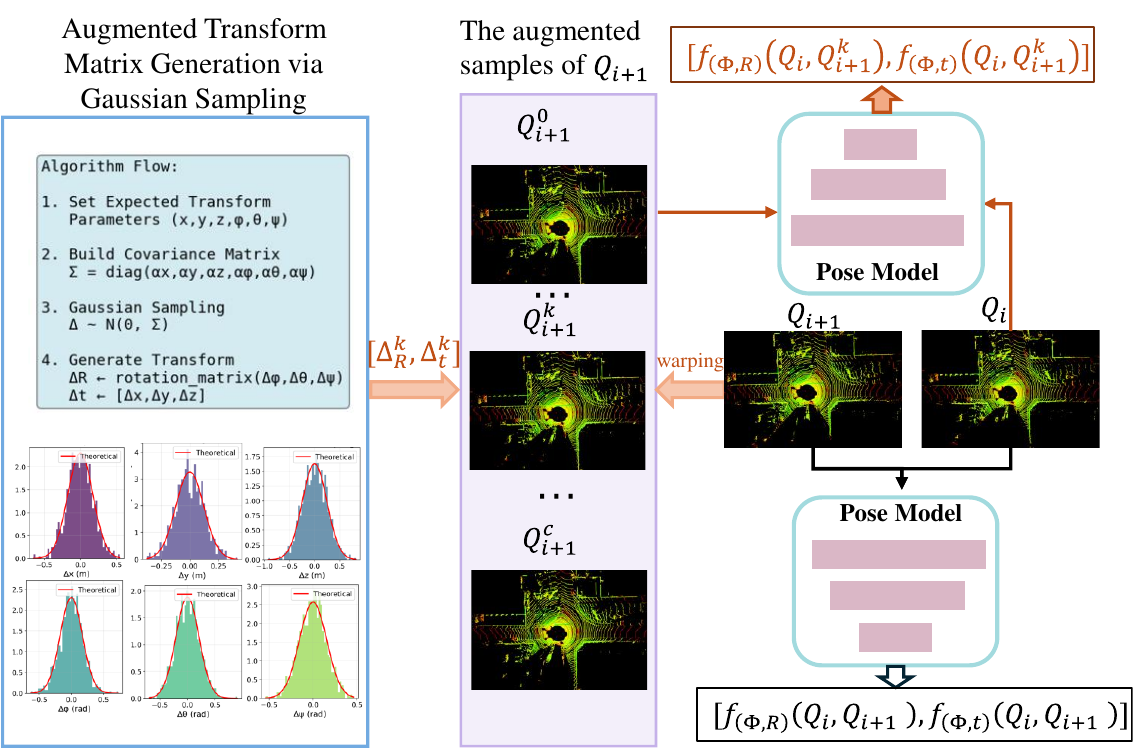}  
		\caption{The details of AIS module: Generation Process of Augmented Samples and Translation Matrix Predictor.}  
\label{delta_Rt_en}  
\vspace{-0.3cm}
\end{figure} 

\subsubsection{Prediction Inconsistency Loss}

We apply augmentations to a sample pair of LiDAR frames to simulate varied motion states. The prediction inconsistency loss is calculated from the variance of poses across augmented pairs, which tests the model's accuracy for varied poses under the same scenes. This loss evaluates the model's reliability and stability of pose regression. Specifically, as shown in~\Cref{delta_Rt_en}, the original pair of LiDAR frames are $\{\mathbf{Q_i}, \mathbf{Q_{i+1}}\}$. Assuming that the six degrees of freedom in the pose are independent, we sample the translations and rotation angles for each axis from a normal distribution with zero means, denoted $\Delta_{\mathbf{R}}^k$ and $\Delta_{\mathbf{t}}^k$. 
We then apply these transformations to $\mathbf{Q_{i+1}}$, producing corresponding augmented samples $\mathbf{Q_{i+1}^k}$ by $\mathbf{Q_{i+1}^k} = \Delta_{\mathbf{R}}^k\mathbf{Q}_{i+1} + \Delta_{\mathbf{t}}^k$,
where $k=0,\cdots,c$. A new sample pair $\{ \mathbf{Q_i}, \mathbf{Q_{i+1}^k} \}$ consisting of the augmented and original samples is fed into the network and outputs the predicted rotation $f_{(\Phi, \mathbf{R})}(\mathbf{Q_i}, \mathbf{Q_{i+1}^k})$ and translation $f_{(\Phi, \mathbf{t})}(\mathbf{Q_i}, \mathbf{Q_{i+1}^k})$, then the rotation matrix $\hat{\mathbf{R}}^k$ and translation matrix $\hat{\mathbf{t}}^k$ between the original sample pair $\{ \mathbf{Q_i}, \mathbf{Q_{i+1}} \}$ are:
% \small
\begin{equation}
\begin{aligned}
\hat{\mathbf{R}}^k = (\Delta_{\mathbf{R}}^k)^{-1}f_{(\Phi, \mathbf{R})}(\mathbf{Q_i}, \mathbf{Q_{i+1}^k}),\\
\hat{\mathbf{t}}^k = (\Delta_{\mathbf{R}}^k)^{-1}(f_{(\Phi, \mathbf{t})}(\mathbf{Q_i}, \mathbf{Q_{i+1}^k})-\Delta_{\mathbf{t}}^k).
\label{cal_Rt}
\end{aligned}
\end{equation}
% \vspace{-0.2cm}
To test the inconsistency of the network's predictions, 
we compute the variance of the relative poses under $c$ augmentations. 
The variance of the translation is calculated as:
% \small
\begin{equation}
\bar{\mathbf{t}} = \frac{1}{c}\sum_{k=1}^c \hat{\mathbf{t}}^{k},
\text{var}(\hat{\mathbf{t}}) = \frac{1}{c} \sum_{k=1}^c \|\hat{\mathbf{t}}^k - \bar{\mathbf{t}}\|_2^2,\\
\end{equation}
where $\bar{\mathbf{t}}$ is the mean of the $c$ augmented translations.

We first convert the rotation matrix $\hat{\mathbf{R}}^{k}$ into Euler angles, then average the Euler angles and reconstruct an average rotation matrix $\bar{\mathbf{R}}$. The distance between the predicted and average rotations is defined by their angular difference, and the variance is
% \small
\begin{equation}
\text{var}(\hat{\mathbf{R}}) = \frac{1}{c} \sum_{k=1}^c [\text{cos}^{-1}(\frac{tr(\bar{\mathbf{R}}^{-1}\hat{\mathbf{R}}^k)-1}{2})]^2,
\label{var_R}
\end{equation}
where $tr()$ is the matrix trace.

To generate reasonable augmented rotations $\Delta_{\mathbf{R}}^k$ and augmented translations $\Delta_{\mathbf{t}}^k$, 
we assume that the 6 degrees of freedom of the pose are independent of each other, and then sample from a normal distribution with a mean of zero to obtain the displacements in each direction $\Delta_x^k,\Delta_y^k,\Delta_z^k$ and the angle of rotation in each axis $\Delta_\phi^k,\Delta_\theta^k,\Delta_\psi^k$:
\begin{equation}
\begin{aligned}
\Sigma = \text{diag}(\alpha x,\alpha y,\alpha z,\alpha\phi,\alpha\theta,\alpha\psi),\\
(\Delta_x^k,\Delta_y^k,\Delta_z^k,\Delta_\phi^k,\Delta_\theta^k,\Delta_\psi^k) \sim \mathcal{N}(\mathbf{0}, \Sigma),\\
\end{aligned}
\end{equation}
where $x,y,z,\phi,\theta,\psi$ is the expected bitmap transformation between this sample pair, and $\alpha$ between $0$ and $1$ is used to control the magnitude of the transformation. Finally $\Delta_{\mathbf{R}}^k$ is chosen as the rotation matrix corresponding to the rotation angles $(\Delta_\phi^k,\Delta_\theta^k,\Delta_\psi^k)$, and $\Delta_{\mathbf{t}}^k$ is chosen as the displacement $(\Delta_x^k,\Delta_y^k,\Delta_z^k)$ the corresponding translation matrix.

Finally, the prediction inconsistency loss is defined as
\begin{equation}
F_{Incon}(\mathbf{Q_i}, \mathbf{Q_{i+1}}) = \text{var}(\hat{\mathbf{R}}) + \text{var}(\hat{\mathbf{t}}).  \label{F_incon} \\
\end{equation}

\paragraph{Overall Loss Function} The overall loss function of the active incremental selection strategy is shown in Eq. (\ref{over_loss}):
% \small
\begin{equation}
    \mathcal{L} = \frac{1}{n}(\alpha \sum_{i=0}^{n-1} F_{Recon}(\mathbf{Q_i}, \mathbf{Q_{i+1}}) + \beta \sum_{i=0}^{n-1} F_{Incon}(\mathbf{Q_i}, \mathbf{Q_{i+1}})), \label{over_loss}
\end{equation}
where $\alpha$ and $\beta$ are hyperparameters, and $n$ represents the total number of frame pairs in the sequence.
During model inference at each iteration, we compute the overall loss function for each sequence in the remaining sample pool.
A higher loss indicates that the sequence is more challenging, making it more important to the model. We prioritize the top $h$ sequences with the highest loss values, incorporating them into the training set pool. This process is repeated iteratively until the model's performance is optimized.

\vspace{-0.3cm}
\subsection{Overall Process of ActiveLO-training}
In summary, Algorithm \ref{alg:training} outlines the complete workflow of our method. The ActiveLO-training process can be summarized the following steps: 1) Use the initial training set selection strategy to choose a subset from the sample pool as the initial training set. 2) Train the model using this initial training set. 3) With the trained model, select additional sequences actively based on the designed scene reconstruction and prediction inconsistency losses. 4) Repeat steps 2) and 3) until the performance is optimized or the predefined maximum number of sequences is reached. The formulas referenced in the algorithm correspond to the formula numbers in the main paper.

\begin{algorithm}[htbp]
% \midrule
\small
	\caption{The active training set selection strategy (ActiveLO training).}
	\begin{algorithmic}
		\STATE 
		\STATE {\textbf{Input:}}
		\STATE {Sample pool sequences \(S_{total}\), sample pool sequences under general weather \(S_{ge}\), the initial training set size \(u\), The network model \(f_{(\Phi, :)}\), the number of sequences added per iteration \(h\), and the number of iterations \(iter\).}

		\STATE {\textbf{Steps:}}
		\STATE {1:} \hspace{0cm} \textbf{if} Using the initial training set selection strategy \textbf{then}
		\STATE {2:} \hspace{0.3cm}  \textbf{for} $s\in S_{ge}$ \textbf{do}
		\STATE {3:} \hspace{0.5cm} Nodes and edges $e_{n_{k-1}{n_k}}$ used to partition the trajectory $n_k$.
            \STATE {4:} \hspace{0.4cm} Calculate the features of the nodes $\theta_{n_{k}}$, $\overline{\theta}$, $\sigma_\theta$ (Eq.(1)\textasciitilde(2).
            \STATE {5:} \hspace{0.4cm} Calculate the features of the edges $l_{n_{k}n_{k+1}}$, $e_{n_{k}n_{k+1}}$, $\overline{v}$, $\sigma_v$, $\overline{l}$, $\sigma_l$ (Eq.(3)\textasciitilde(5)).
            \STATE {6:}  \hspace{0.4cm} Calculate the features of  the samples $s_o$ (Eq.(6)\textasciitilde(7)).
            \STATE {7:}  \hspace{0.4cm} Evaluate the trajectory variability $F_{Var}(s)$ (Eq.(8))) and importance $F_{Impor}(s)$ (Eq.(9)) of each sequence.
            \STATE {8:}  \hspace{0.4cm} Solve the linear programming problem to obtain the selected initial training set $S_{init}$ (Eq.(10).
            \STATE {9:} \hspace{0.3cm}  \textbf{end for}
            \STATE {10:}  \hspace{0cm}  \textbf{end if}
            \STATE{11:}  \hspace{0.3cm}  \textbf{for} $itr\in \{1,2,\dots, iter\}$ \textbf{do}
            \STATE{12:}  \hspace{0.5cm} Train the model $f_{(\Phi, :)}$ using the initial training set $S_{init}$.
            \STATE{13:}  \hspace{0.5cm} Calculate the scene reconstruction loss $F_{Recon}$ (Eq.(11)) for the sequences in the sample pool that were not selected, $s \in S_{total}-S_{init}$.
            \STATE {14:} \hspace{0.5cm} Calculate the prediction inconsistency $F_{Incon}$ (Eq.(16)) for the sequences in the sample pool that were not selected, $s \in S_{total}-S_{init}$.
            \STATE{15:}  \hspace{0.5cm} Calculate the overall loss function (Eq.(17)) for each sequence $s \in S_{total}-S_{init}$ and sort $Sort(s)$.
            \STATE {16:} \hspace{0.5cm} Select the top $h$ sequences from the sorted results as the incremental training set pool $S_{add}$ for this round, and the current training set pool is $S_{selected} = S_{init} \cup S_{add}$.
            \STATE {17:}  \hspace{0.3cm}  \textbf{end for}
		\STATE  {\textbf{Output:}}{Selected training set pool $S_{selected}$}
	\end{algorithmic}
        \label{alg:training}
\end{algorithm}

\vspace{-0.3cm}
\section{Experiment}

\subsection{Experimental Setting}
\label{setting}
\paragraph{Datasets}

We assess our method's performance across diverse outdoor datasets, encompassing general and snowy conditions: KITTI Odometry (KITTI), Ford Campus Vision and Lidar (Ford), Canadian Adverse Driving Conditions (CADC), and Winter Adverse Driving (WADS). We utilize sequences 00-10 from KITTI and over 100m sequences from CADC, totaling 69 sequences for our training sample pool, and select 19 test sequences from Ford, WADS, and CADC.

\paragraph{Implementation Details} 

We initialized our training with 6 sequences from the KITTI Odometry Dataset over 15 epochs, incrementally adding 5 sequences every 5 epochs from the sample pool. Implemented in PyTorch on an NVIDIA GeForce RTX 3090Ti GPU, our model used the Adam optimizer (\(\beta_1 = 0.9\), \(\beta_2 = 0.99\)) with an initial learning rate of 0.001, reduced to 0.0001 post-15 epochs. The backbone network, HPPLO-Net~\cite{zhou2023hpplo}, is a general unsupervised LiDAR odometry method, ideal for assessing training set selection strategies without specialized adverse weather modules.

  \begin{table}[t]
  	\setlength\tabcolsep{2pt}  
    \caption{The overall evaluation results on all test scenes. $t_{rel}$: average translational root mean square error (RMSE) drift (\%); $r_{rel}$: average rotational RMSE drift $(^{\circ}/100m)$. The best performance in each group is highlighted in bold.}
    \centering
    \resizebox{1\linewidth}{!}{
        \begin{tabular}{c|c|cc|cc|cc|cc}
            \toprule[1.3pt]
            \multirow{2}{*}{Percentage} & \multirow{2}{*}{\shortstack{Selection \\ Strategy}}
            & \multicolumn{2}{c|}{dynamic} 
            & \multicolumn{2}{c|}{snowy\_1} & \multicolumn{2}{c|}{snowy\_2} 
            & \multicolumn{2}{c}{Avg} \\ \cline{3-10} 
            & & $t_{rel}$ & $r_{rel}$ & $t_{rel}$ & $r_{rel}$ & $t_{rel}$ & $r_{rel}$ & $t_{rel}$ & $r_{rel}$ \\  
            \toprule[1.3pt]
            $100\%$ & - & \textbf{3.45} & \textbf{1.31} & \textbf{1.12} & \textbf{0.97} & \textbf{0.52} & \textbf{0.71} & \textbf{1.70} & \textbf{1.00} \\ \hline
            $9\%$  & Random & 12.07 & 3.23 & 4.81 & 1.54 & 6.34 & 5.24 & 7.74 & 3.34 \\ 
            $9\%$ & ActiveLO & \textbf{4.26} & \textbf{2.22} & \textbf{1.26} & \textbf{0.81} & \textbf{4.59} & \textbf{3.0} & \textbf{3.37} & \textbf{2.01} \\  \toprule[1.3pt]
            $23\%$  & Random & 4.71 & 2.17 & \textbf{0.95} & 1.53 & 1.05 & 0.92 & 2.24 & 1.54 \\ 
            $23\%$ & ActiveLO & \textbf{3.27} & \textbf{1.73} & 1.10 & \textbf{0.78} & \textbf{0.67} & \textbf{0.86} & \textbf{1.68} & \textbf{1.12} \\  \toprule[1.3pt]
            $38\%$  & Random & 3.74 & 1.58 & 1.20 & 0.89 & 1.18 & 1.53 & 2.04 & 1.33 \\ 
            $38\%$ & ActiveLO & \textbf{3.52} & \textbf{1.23} & \textbf{0.80} & \textbf{0.73} & \textbf{0.64} & \textbf{0.79} & \textbf{1.65} & \textbf{0.92} \\  \toprule[1.3pt]
            $52\%$  & Random & 4.02 & 1.37 & 1.19 & 0.82 & 0.96 & 0.83 & 2.06 & 1.01 \\ 
            $52\%$ & ActiveLO & \textbf{2.69} & \textbf{1.17} & \textbf{0.86} & \textbf{0.73} & \textbf{0.52} & \textbf{0.68} & \textbf{1.36} & \textbf{0.86} \\
            \toprule[1.3pt]  
        \end{tabular}
    }
    \vspace{-0.3cm}
    \label{tb:overall_with_avg}
\end{table}
\subsection{Overall Evaluation Results}
   \begin{figure} [t] 
\centering  
\includegraphics[width=1\linewidth]{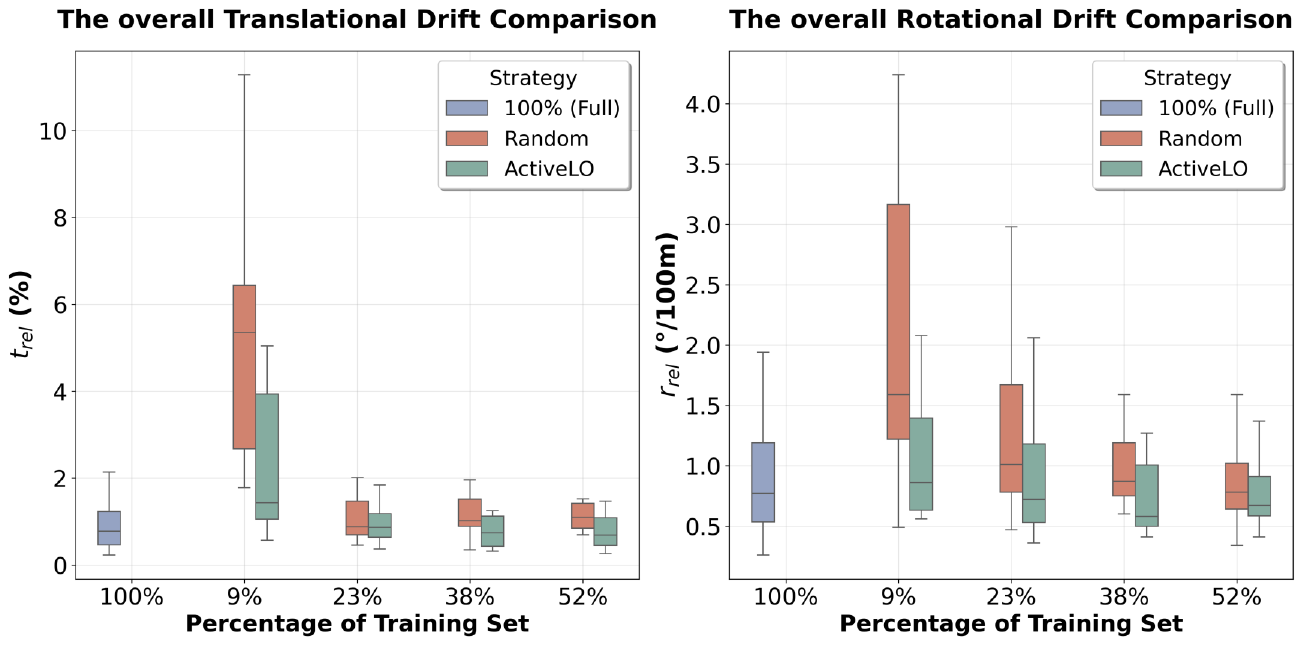}  
		\caption{Overall Distribution of Translation and Rotation Errors for All Test Sequences. ActiveLO consistently and significantly outperforms the random method at every stage, approaching or exceeding 100\% performance with only 52\% of the sequence volume.}  
\label{box_all}  
\vspace{-0.3cm}
\end{figure} 

\begin{table}[t]
 	\setlength\tabcolsep{2pt}  
		\caption{Evaluation results on dynamic scenes. $t_{rel}$: average translational root mean square error (RMSE) drift (\%); $r_{rel}$: average rotational RMSE drift $(^{\circ}/100m)$. The best performance in each group is highlighted in bold.}
    \vspace{-0.2cm}
		\centering
		\resizebox{1\linewidth}{!}{
			\begin{tabular}{c|c|c|cc|cc|cc}
				\toprule[1.3pt]
\multirow{2}{*}{Models} &\multirow{2}{*}{Percentage} &\multirow{2}{*}{\shortstack{Selection \\ Strategy}}
    & \multicolumn{2}{c|}{Ford-1} & \multicolumn{2}{c|}{Ford-2} & \multicolumn{2}{c}{Avg} \\ \cline{4-9} 
&&& $t_{rel}$ & $r_{rel}$ & $t_{rel}$ & $r_{rel}$ & $t_{rel}$ & $r_{rel}$ \\  
\toprule[1.3pt]
\multicolumn{2}{c|}{Full A-LOAM} &-  &\textbf{1.88}&\textbf{0.50} &\textbf{2.05}&\textbf{0.56} &\textbf{1.97} &\textbf{0.53} \\  \hline
    \multicolumn{2}{c|}{A-LOAM} &-  &4.17 &2.00 &{4.72} &1.65  &4.45&1.83 \\ 
   \multicolumn{2}{c|}{ICP-po2po} &- &8.20 &2.64  &16.2 &2.84  &12.2&2.74  \\ 
    \multicolumn{2}{c|}{ICP-po2pl} &-  &3.35 &1.65  &5.68 &1.96   &4.52&1.81 \\ 
    \multicolumn{2}{c|}{VGICP}  &-  &{\textbf{2.51}} &{\textbf{1.03}}  &{\textbf{3.65}} &{\textbf{1.39}} &\textbf{3.08}&\textbf{1.21} \\ \hline
    {Ours} &$100\%$ & -  &\textbf{3.10}	&\textbf{1.39}   &\textbf{3.81} &\textbf{1.22}  &\textbf{3.45}	&\textbf{1.31}
 \\ \hline
    \multirow{2}{*}{Ours}   &$9\%$  & Random &12.85&4.24&11.28&2.21&12.07&3.23 \\ 
     &$9\%$ &ActiveLO &\textbf{3.92}&\textbf{2.84}&\textbf{4.60}&\textbf{1.60}&\textbf{4.26}&\textbf{2.22} \\  \toprule[1.3pt]
    \multirow{2}{*}{Ours}  &$23\%$  & Random &5.59 &2.98 &3.83 &1.36 &4.71 &2.17 \\ 
     &$23\%$ &ActiveLO &\textbf{2.68} &\textbf{2.06} &\textbf{3.86} &\textbf{1.39} &\textbf{3.27} &\textbf{1.73} \\ 
				\toprule[1.3pt]
        \multirow{2}{*}{Ours}   &$38\%$  & Random      &3.90 &1.91 &3.59 &1.25 &3.74 &1.58 \\ 
     &$38\%$ &ActiveLO   &\textbf{3.52} &\textbf{1.27} &\textbf{3.51} &\textbf{1.19} &\textbf{3.52} &\textbf{1.23} \\ 
 	\toprule[1.3pt]
    \multirow{2}{*}{Ours}   &$52\%$  & Random &4.55 &1.59 &3.49 &1.16 &4.02 &1.37 \\ 
     &$52\%$ &ActiveLO &\textbf{2.35} &\textbf{1.28} &\textbf{3.04} &\textbf{1.06} &\textbf{2.69} &\textbf{1.17} \\  	\toprule[1.3pt]
		\end{tabular}}
		\label{tb:ford}
		\vspace{-0.3cm} 
	\end{table}
\begin{figure} [t] 
\centering  
\includegraphics[width=1\linewidth]{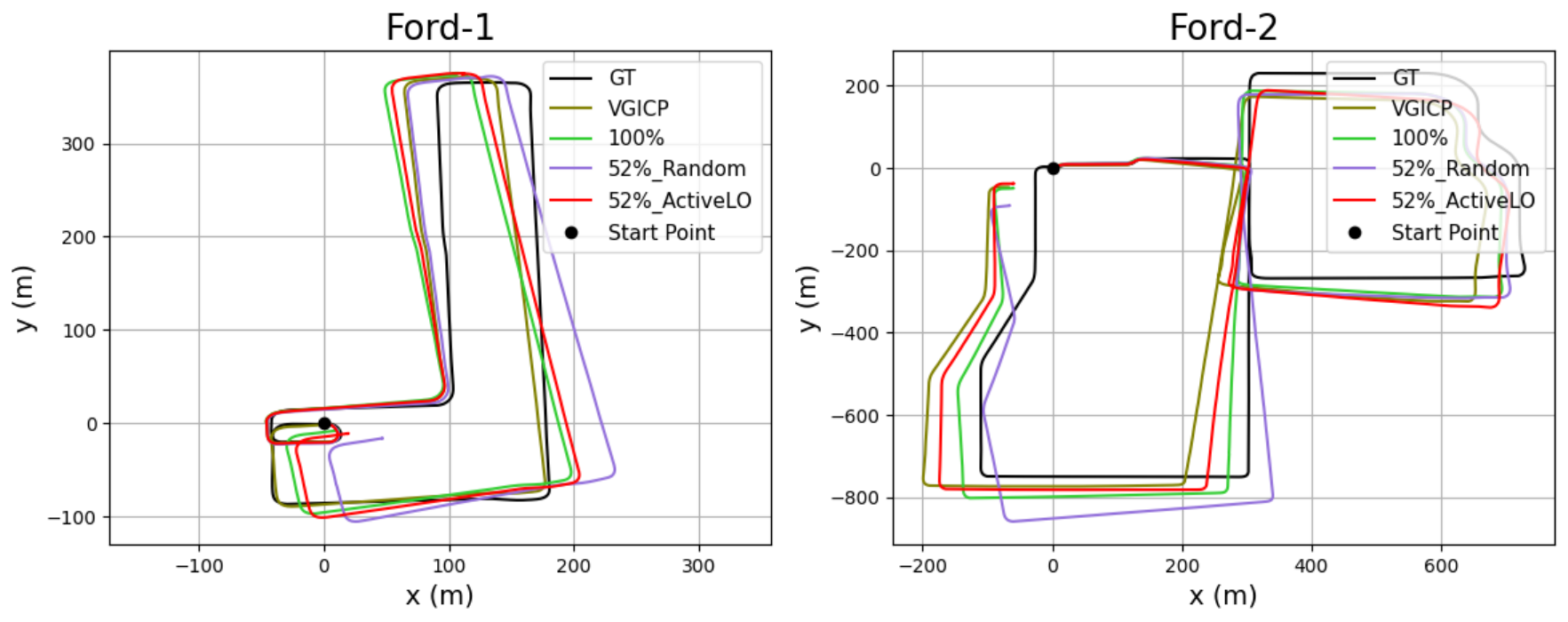}  
		\caption{The trajectory visualization on dynamic sequences. The results indicate our method achieves optimal alignment with ground truth in dynamic scenes.}  
\label{traj_ford}  
\vspace{-0.3cm}
\end{figure}

Our analysis of evaluation results across various scenario (dynamic, snowy\_1, snowy\_2) test sequences, using different training set percentages (Table \ref{tb:overall_with_avg}), demonstrates that ActiveLO-training achieves full-dataset performance with only 52\% of the sequence volume. The 9\% training set, comprising solely the initial training data from KITTI Odometry, provides the model with a baseline pose prediction capability in clear weather. All other training set percentages incorporate the same initial set alongside incremental training data, selected based on different strategies. This efficient data subset selection significantly optimizes computational resources and training time.

Moreover, our method consistently outperforms the random selection strategy, exhibiting stable performance across a multitude of scenarios. This consistency suggests that ActiveLO-training possesses a robust selection mechanism that is less susceptible to the variability inherent in random sampling methods. Figure \ref{box_all} illustrates the overall distribution of rotation and translation errors for all test sequences, clearly demonstrating the superiority of our method in terms of error minimization.
The replicability of our results across multiple experiments further validates the reliability of ActiveLO-training. It consistently identifies the most informative sequences from a vast pool of mixed samples, effectively minimizing the training set size without compromising the model's predictive accuracy. 
This capability is particularly valuable for large and diverse datasets, where selecting a representative subset is crucial for efficient model training.

\begin{table*}[t]
	\setlength\tabcolsep{1.5pt} 
	\caption{The detailed evaluation results on Snowy\_1 (CADC). $t_{rel}$: average translational root mean square error (RMSE) drift (\%); $r_{rel}$: average rotational RMSE drift $(^{\circ}/100m)$. The best performance is in bold.} % Modified caption
	\vspace{-0.2cm}
	\centering
	\resizebox{1\linewidth}{!}{
		\begin{tabular}{c|c|cc|cc|cc|cc|cc|cc|cc|cc|cc|cc}
			\toprule[1.3pt]
			\multirow{2}{*}{Percentage} & \multirow{2}{*}{Slection Strategy}
			& \multicolumn{2}{c|}{41} & \multicolumn{2}{c|}{42} & \multicolumn{2}{c|}{44} & \multicolumn{2}{c|}{45} & \multicolumn{2}{c|}{48} 
			& \multicolumn{2}{c|}{51} & \multicolumn{2}{c|}{54} & \multicolumn{2}{c|}{55} & \multicolumn{2}{c|}{56} 
			& \multicolumn{2}{c}{avg} \\ \cline{3-22}
			&& $t_{rel}$ & $r_{rel}$ & $t_{rel}$ & $r_{rel}$ & $t_{rel}$ & $r_{rel}$
			& $t_{rel}$ & $r_{rel}$ & $t_{rel}$ & $r_{rel}$ & $t_{rel}$ & $r_{rel}$
			& $t_{rel}$ & $r_{rel}$ & $t_{rel}$ & $r_{rel}$ & $t_{rel}$ & $r_{rel}$
			& $t_{rel}$ & $r_{rel}$
			\\
			\toprule[1.3pt]
			\multicolumn{2}{c|}{Full A-LOAM} &\textbf{1.66} &\textbf{0.41} &\textbf{0.31} &\textbf{0.26} &\textbf{0.75} &\textbf{0.20} &\textbf{0.31} &\textbf{0.16} &\textbf{0.93} &\textbf{0.58} &\textbf{0.80} &\textbf{0.27} &\textbf{1.96} &\textbf{0.42} &\textbf{11.97} &\textbf{0.23} &\textbf{18.03} &\textbf{0.39} &\textbf{4.08} &\textbf{0.32} \\ \hline
			\multicolumn{2}{c|}{A-LOAM} &2.70 &0.82 &2.16 &\textbf{0.84} &3.45 &\textbf{0.75} &1.83 &\textbf{1.20} &\textbf{0.99} &\textbf{0.80} &{1.94} &\textbf{0.76} &2.98 &\textbf{0.86} &12.54 &\textbf{0.91} &18.72 &1.80 &5.26 &\textbf{0.97} \\
			\multicolumn{2}{c|}{ICP-po2po} &3.61 &0.96 &4.28 &1.81 &4.35 &2.77 &10.46 &7.92 &6.73 &4.36 &4.17 &2.94 &3.44 &3.22 &6.04 &1.86 &7.28 &2.99 &5.60 &3.20 \\
			\multicolumn{2}{c|}{ICP-po2pl} &4.46 &1.27 &6.98 &1.96 &10.23 &2.98 &6.26 &3.29 &9.72 &4.89 &\textbf{1.23} &2.18 &2.59 &3.71 &3.09 &0.87 &25.93 &1.81 &7.83 &2.55 \\
			\multicolumn{2}{c|}{VGICP} &\textbf{0.72} &\textbf{0.47} &\textbf{1.84} &0.99 &\textbf{2.75} &1.37 &\textbf{1.49} &1.30 &3.06 &3.07 &{2.11} &{2.21} &\textbf{2.06} &{1.76} &\textbf{0.96} &{1.01} &\textbf{1.11} &\textbf{0.74} &\textbf{1.79} &1.44 \\
			\toprule[1pt]
			$100\%$ & - &\textbf{0.23} &\textbf{0.36} &\textbf{0.59} &\textbf{0.56} &\textbf{0.88} &\textbf{0.43} &\textbf{0.43} &\textbf{1.94} &\textbf{0.32} &\textbf{0.77} &\textbf{0.66} &\textbf{0.92} &\textbf{0.41} &\textbf{0.60} &\textbf{0.40} &\textbf{0.26} &\textbf{0.79} &\textbf{0.58} &\textbf{0.52} &\textbf{0.71} \\ \hline
			$9\%$ & Random &6.15 &3.40 &2.47 &0.80 &24.33 &14.19 &6.72 &18.93 &2.88 &1.28 &1.92 &1.19 &1.78 &\textbf{0.49} &8.59 &3.97 &2.22 &2.93 &6.34 &5.24 \\
			$9\%$ &ActiveLO &\textbf{3.95} &\textbf{0.82} &\textbf{2.80} &\textbf{0.81} &\textbf{20.49} &\textbf{10.28} &\textbf{4.52} &\textbf{11.68} &\textbf{1.04} &\textbf{0.86} &\textbf{1.14}&\textbf{0.67}&\textbf{1.64}&0.56&\textbf{5.04}&\textbf{2.08}&\textbf{0.72}&\textbf{1.01} &\textbf{4.59} &\textbf{3.00} \\
			\toprule[1.3pt]
			$23\%$ & Random &0.68 &0.78 &1.58 &0.92 &1.27 &0.98 &1.89 &\textbf{1.89} &0.88 &0.73 &\textbf{0.83} &\textbf{0.78} &\textbf{0.59} &\textbf{0.47} &0.83 &1.06 &0.90 &0.68 &1.05 &0.92 \\
			$23\%$ &ActiveLO &\textbf{0.42}	&\textbf{0.53} &\textbf{0.59}	&\textbf{0.56} &\textbf{0.71}&\textbf{0.72} &\textbf{0.87}&2.65 &\textbf{0.37}	&\textbf{0.67} & 1.19&	1.19 &{0.87} &{0.57} &\textbf{0.50}	&\textbf{0.36} & \textbf{0.55}	&\textbf{0.53} &\textbf{0.67} &\textbf{0.86} \\
			\toprule[1.3pt]
			$38\%$ & Random &0.62 &0.60 &\textbf{0.89} &0.87 &1.15 &0.99 &3.46 &6.49 &0.76 &1.12 &1.38 &1.25 &1.10 &0.76 &0.90 &1.09 &\textbf{0.35} &0.62 &1.18 &1.53 \\
			$38\%$ &ActiveLO &\textbf{0.41} &\textbf{0.51} &1.17 &\textbf{0.64} &\textbf{0.96} &\textbf{0.71} &\textbf{0.42} &\textbf{2.71} &\textbf{0.32} &\textbf{0.42} &\textbf{1.05} &\textbf{0.83} &\textbf{0.52} &\textbf{0.41} &\textbf{0.48} &\textbf{0.43} &{0.44} &\textbf{0.49} &\textbf{0.64} &\textbf{0.79} \\
			\toprule[1.3pt]
			$52\%$ & Random &0.70  &0.53 &1.10 &0.61 &0.95 &\textbf{0.65} &1.12 &1.59 &0.70 &0.85 &1.24 &1.03 &1.03 &0.78 &0.94 &0.86 &0.84 &\textbf{0.60} &0.96 &0.83 \\
			$52\%$ &ActiveLO &\textbf{0.37} &\textbf{0.41} &\textbf{0.69} &\textbf{0.61} &\textbf{0.72} &0.88 &\textbf{0.48} &\textbf{1.02} &\textbf{0.26} &\textbf{0.58} &\textbf{0.98} &\textbf{0.94} &\textbf{0.48} &\textbf{0.49} &\textbf{0.42} &\textbf{0.56} &\textbf{0.31} &0.67 &\textbf{0.52} &\textbf{0.68} \\
			\toprule[1.3pt]
		\end{tabular}
	}
	\label{tb:cadc}
	\vspace{-0.3cm}
\end{table*}
% \vspace{-0.2cm}

\begin{table*}[t]
	\setlength\tabcolsep{3pt}  
	\caption{The detailed evaluation results on snowy\_2 (WADS). $t_{rel}$: average translational root mean square error (RMSE) drift (\%); $r_{rel}$: average rotational RMSE drift $(^{\circ}/100m)$. The best performance is in bold.}
	\centering
	\resizebox{1\linewidth}{!}{
	\begin{tabular}{c|c|cc|cc|cc|cc|cc|cc|cc|cc|cc}
		\toprule[1.3pt]
		\multirow{2}{*}{Percentage} & \multirow{2}{*}{Section strategy} & \multicolumn{2}{c|}{13} & \multicolumn{2}{c|}{14} & \multicolumn{2}{c|}{16} & \multicolumn{2}{c|}{20} & \multicolumn{2}{c|}{34} & \multicolumn{2}{c|}{35} & \multicolumn{2}{c|}{36} & \multicolumn{2}{c|}{37} & \multicolumn{2}{c}{avg} \\
		\cline{3-20}
		& & $t_{rel}$ & $r_{rel}$ & $t_{rel}$ & $r_{rel}$ & $t_{rel}$ & $r_{rel}$ & $t_{rel}$ & $r_{rel}$ & $t_{rel}$ & $r_{rel}$ & $t_{rel}$ & $r_{rel}$ & $t_{rel}$ & $r_{rel}$ & $t_{rel}$ & $r_{rel}$ & $t_{rel}$ & $r_{rel}$ \\
		\toprule[1.3pt]
		\multicolumn{2}{c|}{Full A-LOAM} & \textbf{0.96} & \textbf{0.51} & \textbf{0.99} & \textbf{0.90} & \textbf{1.16} & \textbf{0.83} & \textbf{1.72} & \textbf{0.82} & \textbf{1.05} & \textbf{0.43} & \textbf{1.16} & \textbf{0.43} & \textbf{0.74} & \textbf{0.51} & \textbf{0.84} & \textbf{0.85} & \textbf{1.20} & \textbf{0.66} \\ \hline
		\multicolumn{2}{c|}{A-LOAM} & 14.09 & 2.20 & 5.44 & \textbf{1.02} & 6.94 & 2.06 & 13.55 & 1.46 & 3.70 & \textbf{0.61} & 5.51 & \textbf{0.81} & 9.42 & 2.13 & 11.90 & 4.99 & 8.82 & 1.91 \\
		\multicolumn{2}{c|}{ICP-po2po} & 1.70 & 1.51 & 1.64 & 2.01 & 1.48 & 1.93 & 2.04 & 1.22 & 1.11 & 0.62 & 1.27 & 0.85 & 0.92 & 0.90 & 0.73 & 0.74 & 1.36 & 1.22 \\
		\multicolumn{2}{c|}{ICP-po2pl} & \textbf{0.83} & \textbf{0.31} & 1.72 & 1.99 & \textbf{0.97} & \textbf{1.07} & \textbf{0.75} & \textbf{0.70} & 2.11 & 1.36 & \textbf{1.08} & 0.99 & 0.79 & 0.98 & \textbf{0.52} & \textbf{0.72} & \textbf{1.10} & 1.02 \\
		\multicolumn{2}{c|}{VGICP} & 1.21 & 0.76 & \textbf{1.01} & 1.15 & 1.27 & 1.10 & 1.73 & 0.96 & \textbf{1.04} & 0.98 & 1.20 & 0.89 & \textbf{0.78} & \textbf{0.57} & 0.77 & 1.13 &{1.13} &\textbf{0.94} \\
		\toprule[1pt]
		$100\%$ & - & \textbf{0.97} & \textbf{0.51} & \textbf{0.50} & \textbf{1.23} & \textbf{1.17} & \textbf{1.84} & \textbf{2.14} & \textbf{1.03} & \textbf{1.44} & \textbf{1.16} & \textbf{1.29} & \textbf{0.36} & \textbf{0.78} & \textbf{0.57} & \textbf{0.64} & \textbf{1.07} & \textbf{1.12} & \textbf{0.97} \\ \toprule[1pt]
		$9\%$ & Random  &5.71 &1.52 &4.73 &1.16 &5.88 &1.21 &\textbf{2.29} &2.64 &5.35 &1.23 &5.53 &1.59 &4.25 &1.27 &4.70 &1.73 &4.81 &1.54 \\
		$9\%$ & ActiveLO & \textbf{1.43} & \textbf{1.18} & \textbf{0.77} & \textbf{0.62} & \textbf{1.12} & \textbf{1.08} & {3.24} & \textbf{1.19} & \textbf{1.09} & \textbf{0.61} & \textbf{1.07} & \textbf{0.64} & \textbf{0.77} & \textbf{0.57} & \textbf{0.57} & \textbf{0.58} & \textbf{1.26} & \textbf{0.81} \\ \toprule[1pt]
		$23\%$ & Random &\textbf{0.88} &0.77 &\textbf{0.46} &1.51 &\textbf{0.84} &2.12 &2.01 &2.50 &1.36 &1.83 &\textbf{0.65} &1.01 &0.70 &0.98 &\textbf{0.69} &1.49 &\textbf{0.95} &1.53 \\
		$23\%$ & ActiveLO & {1.00} & \textbf{0.45} & {0.71} & \textbf{0.36} & {1.38} & \textbf{1.25} & \textbf{{1.84}} & \textbf{1.17} & \textbf{1.16} & \textbf{0.80} & {1.15} & \textbf{0.91} & \textbf{0.69} & \textbf{0.48} & 0.84 & \textbf{0.82} & 1.10 & \textbf{0.78} \\ \toprule[1pt]
		$38\%$ & Random  &0.98 &\textbf{0.67} &0.86 &0.75 &1.64 &1.59 &1.96 &\textbf{1.13} &1.33 &0.79 &1.02 &0.62 &0.91 &0.84 &0.91 &0.75 &1.20 &0.89 \\
		$38\%$ & ActiveLO & \textbf{0.85} & 0.69 & \textbf{0.34} & \textbf{0.50} & \textbf{1.15} & \textbf{1.23} & \textbf{1.25} & 1.18 & \textbf{1.10} & \textbf{0.50} & \textbf{0.74} & \textbf{0.57} & \textbf{0.60} & \textbf{0.58} & \textbf{0.40} & \textbf{0.57} & \textbf{0.80} & \textbf{0.73} \\ \toprule[1pt]
		$52\%$ & Random  &1.41 &\textbf{0.34} &0.77 &\textbf{0.63} &1.52 &1.58 &\textbf{1.43} &1.01 &1.30 &0.75 &0.78 &0.66 &1.42 &0.74 &0.86 &0.81 & 1.19 &0.82 \\
		$52\%$ & ActiveLO 
        & \textbf{0.90} & {0.45} & \textbf{0.33} &{0.67} &\textbf{ 1.20} & \textbf{1.37} & {1.47} & \textbf{0.72} & \textbf{1.21} & \textbf{0.59} & \textbf{0.69} & \textbf{0.65} & \textbf{0.60} & \textbf{0.70} & \textbf{0.49} & \textbf{0.70} & \textbf{0.86} & \textbf{0.73} \\
		\bottomrule[1.3pt]
	\end{tabular}}
	\label{tb:WADS}
\end{table*}

   \begin{figure*} [t] 
\centering  
\includegraphics[width=1\linewidth]{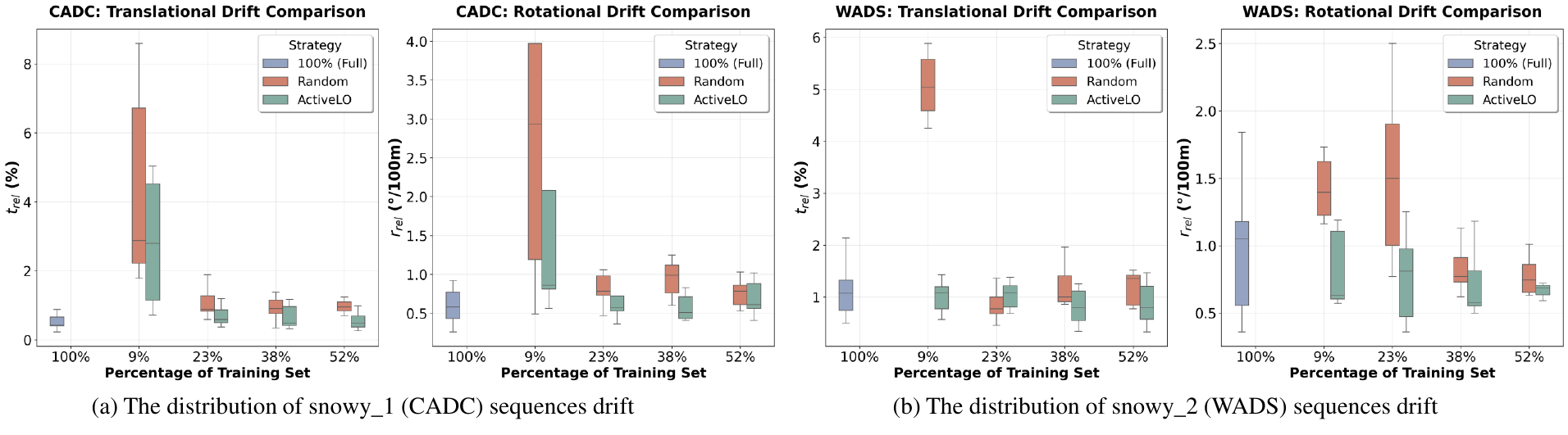}  
		\caption{The drift distribution of snowy sequences.}  
\label{box_overall}  
% \vspace{-0.2cm}
\end{figure*}

   \begin{figure*} [t] 
\centering  
\includegraphics[width=0.95\linewidth]{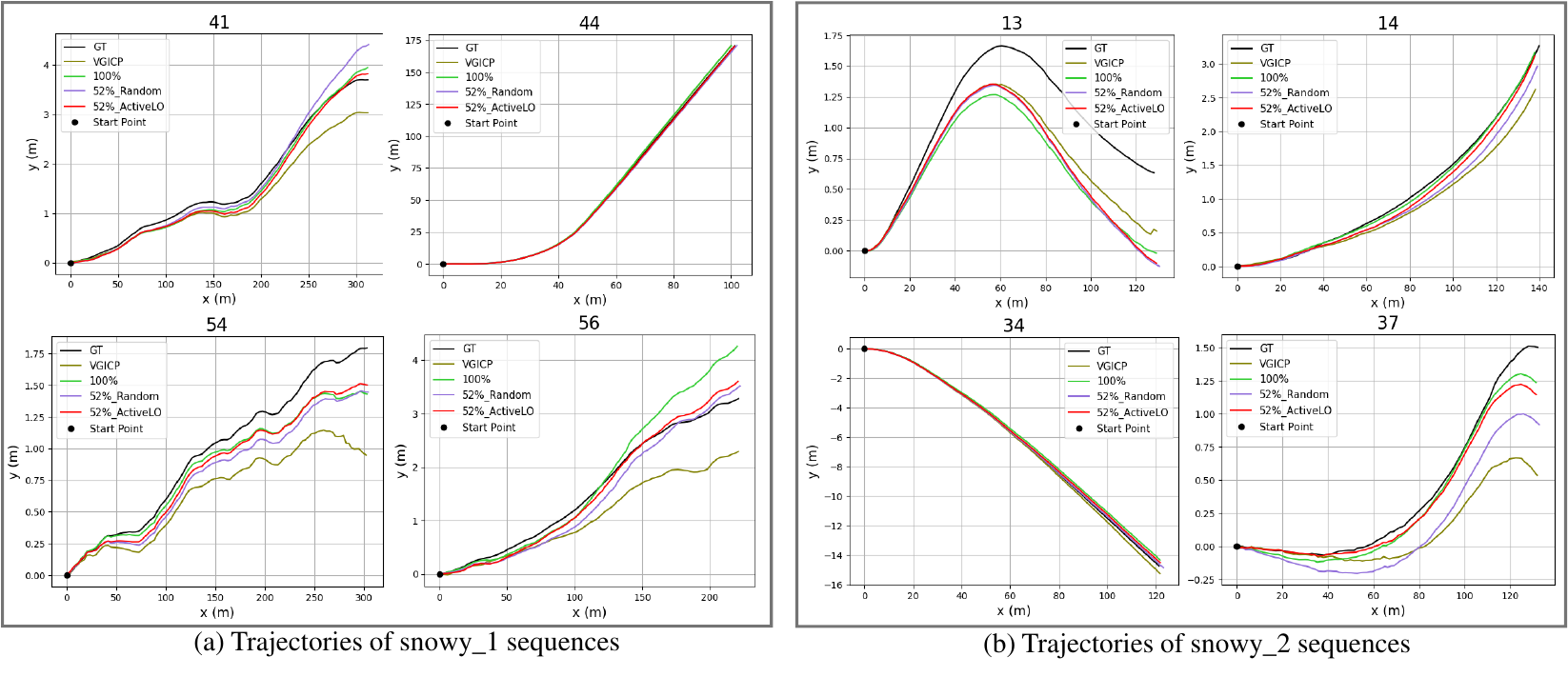}  
		\caption{The trajectory visualization on snowy sequences.}  
\label{traj_snowy}  
% \vspace{-0.2cm}
\end{figure*}
 
\subsection{Evaluation Results on dynamic scenes}

To assess the model's generalization capability across diverse scenarios, we conduct evaluations under clear weather conditions with highly dynamic environments, which degrade LiDAR odometry performance.
A comprehensive comparative analysis on the Ford dataset (Table~\ref{tb:ford}) benchmarks our ActiveLO training framework against random selection baselines and classical LiDAR odometry (LO) methods, including: Full A-LOAM~\cite{aloam}, A-LOAM~\cite{aloam}, ICP-po2po~\cite{ICP}, ICP-po2pl~\cite{ICP}, and VGICP~\cite{koide2021voxelized}.
Notably, our method, trained on 52\% of the sequence volume, outperforms all but Full A-LOAM and even surpasses full-dataset-trained (100\%) models. However, from the following experiments, we know that Full A-LOAM isn't able to maintain a stable superiority across all scenarios. This highlights ActiveLO's efficiency and generalization capabilities.
Figure~\ref{traj_ford} shows the Ford sequence trajectories, where 52\%-trained ActiveLO achieves the closest alignment with the ground-truth trajectories. Consequently, our method demonstrates superior performance with reduced drift and enhanced consistency in dynamic scenes, corroborating Table~\ref{tb:ford}'s quantitative results.

\subsection{Evaluation results on snowy scene (CADC)}
\label{cadc_evaluation}
To evaluate the performance of our ActiveLO-training strategy in snowy scenes, we selected 9 sequences as our test set from the CADC dataset, which is not included in training set, focusing on dates 2018\_03\_06, 2018\_03\_07, and 2019\_02\_27. Table~\ref{tb:cadc} summarizes the evaluation results. It can be seen that ActiveLO-training outperforms random selection, with a 52\% subset surpassing all traditional methods listed.
Despite occasional superior results from random selection due to the CADC's high sequence prevalence, ActiveLO maintains stable performance across datasets. Consistently, ActiveLO-training targets challenging scenarios like heavy snowfall, dynamic objects, and high velocities. The most challenging sequences, such as 0050, 0054, and 0078 from 2019\_02\_27, are always prioritized.
This focus results in a significant outlier proportion, impacting the network's relative pose prediction accuracy. Sequences from 2018\_03\_06, with rich structures and minimal noise, are often excluded. Integrating a diverse range of sequences into the training set reduces errors across test datasets, confirming the effectiveness of ActiveLO-training in enhancing the model's predictive accuracy.
 
  % \vspace{-0.2cm}
\subsection{Evaluation Results on snowy scene (WADS)}
To further test the validity of the model in snowfall conditions,
we selected 8 sequences from the WADS dataset as test sequences. This dataset feature varying degrees of real snowfall and were not present in the training set, allowing their evaluation results to better reflect the impact of different selection strategies on generalization.

As indicated in Table~\ref{tb:WADS}, ActiveLO-training achieves superior performance compared to traditional methods in both translation and rotation error when utilizing only 52\% of the sequences. Notably, it significantly outperforms random selection at the 9\%, 38\%, and 52\% sequence levels, and even surpasses the results obtained using 100\% of the training data. 
The inherent stochasticity of the Random strategy can introduce bias during training, potentially favoring specific scenarios. This occasional bias is evidenced by instances where the Random strategy outperformed the ActiveLO strategy in approximately 23\% of the observed results for certain sequences.
Furthermore, within the WADS dataset specifically, 38\% of the ActiveLO results exhibited a slight superiority over the 52\% of the ActiveLO strategy for the same percentage of sequences. However, a more comprehensive analysis across multiple datasets (Table~\ref{tb:overall_with_avg}) reveals that the ActiveLO strategy demonstrates superior stability in 52\% of cases, underscoring its generalizable robustness despite occasional localized fluctuations.
 Collectively, these results underscore the necessity of a systematic selection strategy to ensure consistent performance across diverse datasets.

  % \vspace{-0.2cm}
\subsection{Ablation Study}
We conduct ablation studies to evaluate the effectiveness of each component in our method, separately analyzing the initial training set selection strategy (ITSS) and the active incremental selection (AIS) strategy.

\paragraph{Initial Training Set Selection}

To assess ITSS's efficacy, we select four sequences on KITTI Odometry  (00, 01, 02, and 08) by ITSS as the initial training set, as outlined in Table \ref{ab:kitti}. To control for randomness, we performed three random selections: Random\_1, Random\_2, and Random\_3, ensuring a balanced comparison.
Following 50 epochs of training, the model was evaluated on the KITTI Odometry and Ford datasets. The results indicate that our model outperforms randomly trained models, especially in generalizing to unseen scenarios (Ford dataset), demonstrating robustness and effectiveness.

 \begin{table}[t]
 	\setlength\tabcolsep{3pt} 
		\caption{Ablation study of ITSS. $t_{rel}$: average translational root mean square error (RMSE) drift (\%); $r_{rel}$: average rotational RMSE drift $(^{\circ}/100m)$. The best performance is in bold.}
    \vspace{-0.2cm}
		\centering
		\resizebox{0.8\linewidth}{!}{
			\begin{tabular}{c|c|cc|cc}
				\toprule[1.3pt]
\multirow{2}{*}{Methods} &\multirow{2}{*}{Training Set} 
  & \multicolumn{2}{c|}{KITTI}   & \multicolumn{2}{c}{Ford}  \\ \cline{3-6} 
&& $t_{rel}$ & $r_{rel}$ & $t_{rel}$ & $r_{rel}$  \\  \hline
 Random1    &01, 04, 05, 09 &6.19 &2.29 &5.27 &2.11 \\  \hline
Random2  &02, 06, 08, 10 &6.28 &1.96 &5.93 &1.92 \\  	\hline
Random3   &03, 05, 07, 10 &8.12 &2.20 &4.03 &1.59 \\  \hline
 Ours    &00, 01, 02, 08 &\textbf{5.31 }  &\textbf{1.89} &\textbf{3.51} &\textbf{1.43}  \\  	\toprule[1.3pt]
		\end{tabular}}
		\label{ab:kitti}
		% \vspace{-0.3cm} 
	\end{table}
    
\begin{table}[t]
	\setlength\tabcolsep{2pt}  
		\caption{Ablation study of AIS. $t_{rel}$: average translational root mean square error (RMSE) drift (\%); $r_{rel}$: average rotational RMSE drift $(^{\circ}/100m)$. The best performance is in bold.}
    \vspace{-0.2cm}
		\centering
		\resizebox{1\linewidth}{!}{
			\begin{tabular}{c|c|cc|cc|cc|cc}
				\toprule[1.3pt]
\multirow{2}{*}{Percentage} &\multirow{2}{*}{Methods} 
    & \multicolumn{2}{c|}{dynamic} & \multicolumn{2}{c|}{snowy\_1} & \multicolumn{2}{c|}{snowy\_2} & \multicolumn{2}{c}{Avg} \\ \cline{3-10} 
&& $t_{rel}$ & $r_{rel}$ & $t_{rel}$ & $r_{rel}$ & $t_{rel}$ & $r_{rel}$  & $t_{rel}$ & $r_{rel}$  \\  
\toprule[1.3pt]
    \multirow{3}{*}{$23\%$}   &SRL  &5.50&1.66 &2.00 &1.45 & 1.62 &1.04 & 3.04 & 1.38 \\ 
     &PIL  &5.79&1.99 &1.73 &1.84 &0.83 &1.43  & 2.78 & 1.75 \\  
     &SRL+PIL & \textbf{3.27} & \textbf{1.73} & \textbf{1.10} & \textbf{0.78} & \textbf{0.67} & \textbf{0.86} & \textbf{1.68} & \textbf{1.12}\\
     \toprule[1.3pt]
        \multirow{3}{*}{$38\%$}  
       &SRL  &5.18 &1.86  & 1.87 &1.38 &1.13 &1.42 & 2.73 & 1.55 \\ 
     &PIL &4.64 &1.59  &1.28 &0.93 &0.84 &1.00 & 2.25 & 1.17  \\  
     &SRL+PIL & \textbf{3.52} & \textbf{1.23} & \textbf{0.80} & \textbf{0.73} & \textbf{0.64} & \textbf{0.79} & \textbf{1.65} & \textbf{0.92}\\	\toprule[1.3pt]
    \multirow{3}{*}{$52\%$}   
    &SRL &3.45 &1.44  &0.99 &0.81 &0.95 &0.91 & 1.80 & 1.05\\ 
     &PIL  &2.91 &1.30 &1.07 &0.77 &0.74 &0.90 & 1.57 & 0.99 \\  
     &SRL+PIL & \textbf{2.69} & \textbf{1.17} & \textbf{0.86} & \textbf{0.73} & \textbf{0.52} & \textbf{0.68} & \textbf{1.36} & \textbf{0.86}  \\  	\toprule[1.3pt]
		\end{tabular}}
		\label{tb:Ab2}
		% \vspace{-0.5cm} 
	\end{table}

\paragraph{Incremental Selection Strategy}

We assessed model performance using three training approaches: Scene Reconstruction Loss (SRL), Prediction Inconsistency Loss (PIL), and a combination of both (SRL+PIL). Table \ref{tb:Ab2} compares these methods across different sequence volumes. For all three approaches, we employed the same training strategy: an initial model was trained on 9\% of the sequences, followed by the addition of the 5 sequences with the highest metrics every 5 epochs. Since all methods shared the same 9\% initial training set, our comparison of incremental selection begins at 23\%. While both SRL and PIL alone yielded better performance individually, their combination provided the most robust model performance across multiple datasets simultaneously.

\begin{figure} [t] 
\centering  
\includegraphics[width=1\linewidth]{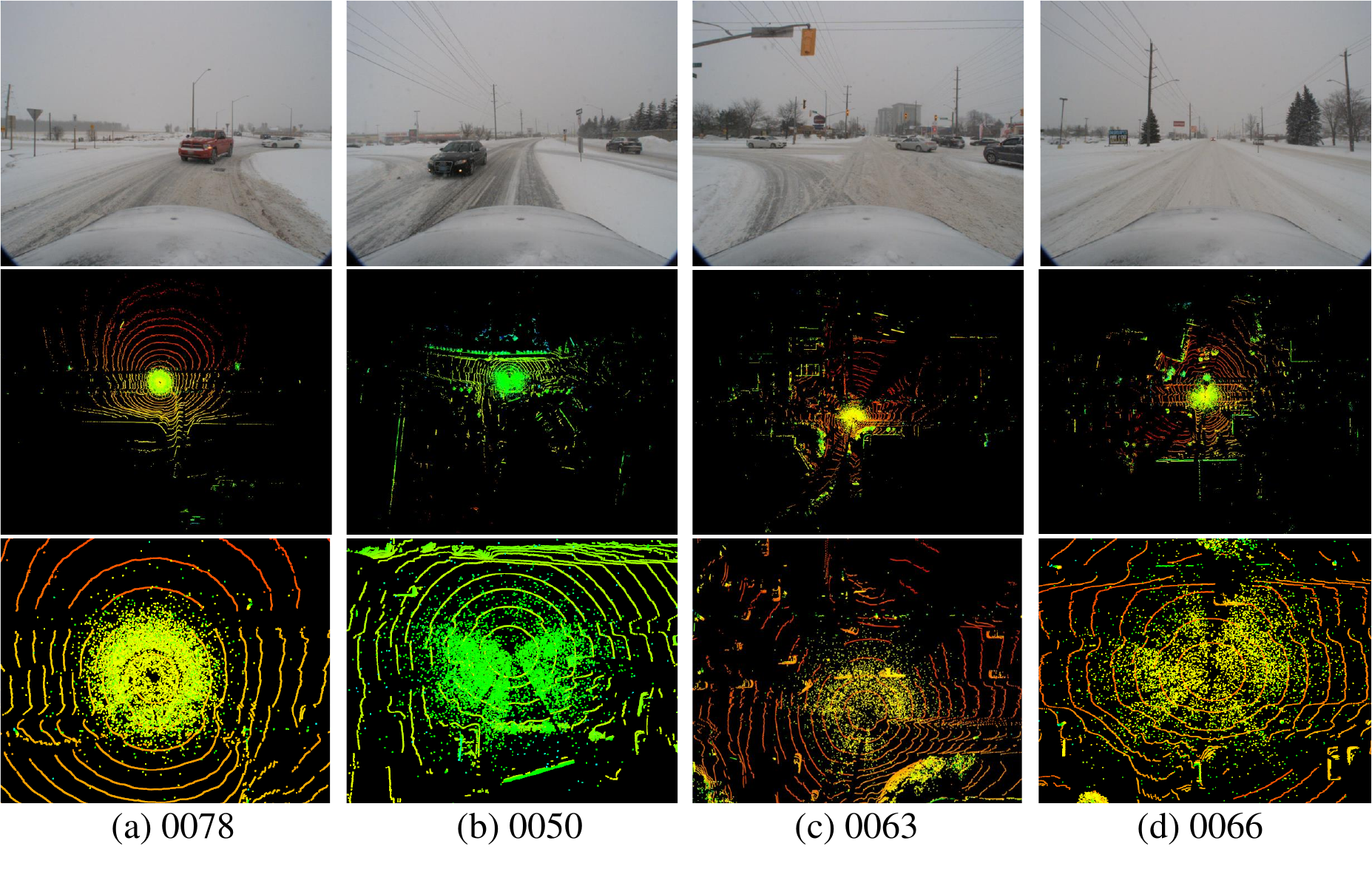}  
    \vspace{-0.5cm}
  \caption{Scene visualization. (a)(b): Scenes extreme open with blizzards and dynamic objects; (c): Scenes with heavy snow and dynamic objects; (d): Scenes with heavy snow and open areas.}
\label{vis_scene}  
% \vspace{-0.2cm}
\end{figure}

The Scene Reconstruction Loss (SRL) metric primarily assesses a model's predictive capability based on scene reconstruction, while the Prediction Inconsistency Loss (PIL) metric reflects this capability through inconsistencies in transformation matrices.  
Our comparative analysis of their incremental sequence selection process revealed that roughly 80\% of the sequences in the final training set pool were identical, even with minor differences in their selection paths. Initially, the model struggled with abrupt scene changes, such as those from clear to snowy conditions, leading to poor performance in most snowy environments. However, after 3-4 rounds of incremental selection, the model progressively adapted to these snowy scenes. We observed that in later stages (38\%-52\% of incremental selection), SRL tended to prioritize sequences with heavy snowfall and numerous dynamic objects, whereas PIL favored sequences with sparser scenes. This divergence occurs because once the network's predictive capabilities stabilize, scenes with heavy snow and dynamic objects introduce more noise, resulting in higher SRL values. Conversely, sparse scene point clouds are less dense, making it difficult to accurately capture small augmentation transformations, which leads to higher PIL values. Both types of scenes represent relatively challenging sequences within the sample pool. Therefore, these two metrics complement each other, collaboratively selecting highly important sequences for inclusion in the training set. Notably, the most challenging samples—those combining heavy snow, sparse environments, and dynamic objects—were preferentially selected by both metrics in the earlier stages.
Figure \ref{vis_scene} displays four scene examples, where the most challenging sequences (a) and (b) are early selections by both metrics. Sequences like (c) are later selected by SRL, and (d) by PIL.

  % \vspace{-0.2cm}
 \subsection{Model Training Efficiency}

 % \vspace{-0.2cm}
\begin{table}[h]
    \caption{Comparison of training efficiency.}	
    \centering
    \resizebox{0.85\linewidth}{!}{
        \begin{tabular}{c||c|c|c|c} \toprule[1.5pt]
            {Module} &epochs & {Training} &Inference &Total  \\ \hline
            100\%  &50 &3450 &- &3450 \\
            ActiveLO &50 & 1000 &336 &\textbf{1336}  \\ \hline
    \end{tabular}}
    \label{tab:time}
    % \vspace{-0.3cm}	
\end{table}
Table~\ref{tab:time} shows the comparison of training efficiency with the full sample pool (100\%) and our ActiveLO-training. 
For the full sample pool (100\%), assuming the total number of sequences is \( Num(S_{total}) \), at least 50 training epochs are needed for the model to converge to its optimal state. Therefore, the total number of training sequence iterations is
 \begin{equation}
    L_{full} = Num(S_{total}) \times 50. 
 \end{equation}

In contrast, ActiveLO-training involves model training (both initial and incremental stages) and model inference, the total number of training sequence iterations is given by:
  \begin{equation}
     L_{52}^{train} = Num(S_{init}) \times 15 + \sum_{itr=1}^{iter+1}(Num(S_{init})+h\times itr)\times 5. 
 \end{equation}
  Here, \( Num(S_{init}) \) represents the number of sequences in the initial training set, \( h \) is the number of sequences added in each round, and \( iter \) is the number of iterations required to reach 52\% of the total sequences. 
However, it is uncertain whether the model will reach peak performance during training stage, 
so we iterate another round ($iter+1$). If the new round's evaluation results no longer improve, we stop the training. 
Moreover,  ActiveLO-training requires evaluating the loss function metrics for the remaining sequences in the sample pool before each round of incremental additions, this introduces additional training overhead compared to training with the full sample pool. The calculation process for this additional overhead is as follows:
% \small
  \begin{equation}
  \begin{aligned}
     L_{52}^{remain} = Num(S_{total})-Num(S_{init}) + \\ \sum_{itr=1}^{iter}(Num(S_{total})-Num(S_{init})-h\times itr).
\end{aligned}
 \end{equation}
In our experiments, \( Num(S_{total}) \), \( Num(S_{init}) \), \( h \), and \( iter \) are 69, 6, 5, and 7, respectively. The calculated training overhead is presented in Table \ref{tab:time}. We can see that our ActiveLO-training strategy significantly improves training efficiency compared to using the full sample pool (100\%).

\section{Conclusion}
This paper presents an ActiveLO-training strategy for deep learning-based LiDAR Odometry (LO), designed to optimize training set selection for improved efficiency and model generalization across diverse weather conditions. Our approach uses Initial Training Set Selection (ITSS) to measure sequence complexity and motion variability, assembling a diverse initial training set through sequence diversity metrics. For challenging snowy sequences, we apply an Active Incremental Selection (AIS) that actively incorporates key sequences based on Scene Reconstruction Loss (SRL) and Prediction Inconsistency Loss (PIL) metrics. This iterative process progressively refines the training set until optimal model performance is achieved. Empirical results across multiple datasets show that our method achieves performance comparable to models trained on the entire sample pool, while significantly improving training efficiency. Future work will explore segmenting sequences within the sample pool to further enhance training efficiency, and add data from more diverse scenarios to the sample pool.

\bibliography{main}

\begin{thebibliography}{10}
\providecommand{\url}[1]{#1}
\csname url@samestyle\endcsname
\providecommand{\newblock}{\relax}
\providecommand{\bibinfo}[2]{#2}
\providecommand{\BIBentrySTDinterwordspacing}{\spaceskip=0pt\relax}
\providecommand{\BIBentryALTinterwordstretchfactor}{4}
\providecommand{\BIBentryALTinterwordspacing}{\spaceskip=\fontdimen2\font plus
\BIBentryALTinterwordstretchfactor\fontdimen3\font minus \fontdimen4\font\relax}
\providecommand{\BIBforeignlanguage}[2]{{%
\expandafter\ifx\csname l@#1\endcsname\relax
\typeout{** WARNING: IEEEtran.bst: No hyphenation pattern has been}%
\typeout{** loaded for the language `#1'. Using the pattern for}%
\typeout{** the default language instead.}%
\else
\language=\csname l@#1\endcsname
\fi
#2}}
\providecommand{\BIBdecl}{\relax}
\BIBdecl

\bibitem{Matsuki_2024_CVPR}
H.~Matsuki, R.~Murai, P.~H. Kelly, and A.~J. Davison, ``Gaussian splatting slam,'' in \emph{IEEE Conf. Comput. Vis. Pattern Recog.}, pp. 18\,039--18\,048.

\bibitem{10681669}
Z.~Rui, W.~Xu, and Z.~Feng, ``Fast-lidar-slam: A robust and real-time factor graph for urban scenarios with unstable gps signals,'' \emph{IEEE Trans. Intell. Transport. Syst.}, pp. 20\,043--20\,058, 2024.

\bibitem{10938051}
F.~Li, C.~Fu, J.~Wang, and D.~Sun, ``Dynamic semantic slam based on panoramic camera and lidar fusion for autonomous driving,'' \emph{IEEE Trans. Intell. Transport. Syst.}, pp. 1--14, 2025.

\bibitem{zhang2023perception}
Y.~Zhang, A.~Carballo, H.~Yang, and K.~Takeda, ``Perception and sensing for autonomous vehicles under adverse weather conditions: A survey,'' \emph{ISPRS J. Pho. Remote Sen.}, pp. 146--177, 2023.

\bibitem{carballo2020libre}
A.~Carballo, J.~Lambert, A.~Monrroy, D.~Wong, P.~Narksri, Y.~Kitsukawa, E.~Takeuchi, S.~Kato, and K.~Takeda, ``Libre: The multiple 3d lidar dataset,'' in \emph{IEEE Intelligent Veh. Symp.}\hskip 1em plus 0.5em minus 0.4em\relax IEEE, 2020, pp. 1094--1101.

\bibitem{mohammed2020perception}
A.~S. Mohammed, A.~Amamou, F.~K. Ayevide, S.~Kelouwani, K.~Agbossou, and N.~Zioui, ``The perception system of intelligent ground vehicles in all weather conditions: A systematic literature review,'' \emph{Sensors}, 2020.

\bibitem{gupta2024robust}
H.~Gupta, O.~Kotlyar, H.~Andreasson, and A.~J. Lilienthal, ``Robust object detection in challenging weather conditions,'' in \emph{Winter Conf. Appl. Comput. Vis.}, 2024, pp. 7523--7532.

\bibitem{jia2022multi}
X.~Jia, L.~Sun, H.~Zhao, M.~Tomizuka, and W.~Zhan, ``Multi-agent trajectory prediction by combining egocentric and allocentric views,'' in \emph{Conference on Robot Learning}.\hskip 1em plus 0.5em minus 0.4em\relax PMLR, 2022, pp. 1434--1443.

\bibitem{jia2023hdgt}
X.~Jia, P.~Wu, L.~Chen, Y.~Liu, H.~Li, and J.~Yan, ``Hdgt: Heterogeneous driving graph transformer for multi-agent trajectory prediction via scene encoding,'' \emph{IEEE Trans. Pattern Anal. Mach. Intell.}, 2023.

\bibitem{shi2022motion}
S.~Shi, L.~Jiang, D.~Dai, and B.~Schiele, ``Motion transformer with global intention localization and local movement refinement,'' \emph{Adv. Neural Inform. Process. Syst.}, pp. 6531--6543, 2022.

\bibitem{vora2020pointpainting}
S.~Vora, A.~H. Lang, B.~Helou, and O.~Beijbom, ``Pointpainting: Sequential fusion for 3d object detection,'' in \emph{IEEE Conf. Comput. Vis. Pattern Recog.}, 2020, pp. 4604--4612.

\bibitem{lu2024activead}
H.~Lu, X.~Jia, Y.~Xie, W.~Liao, X.~Yang, and J.~Yan, ``Activead: Planning-oriented active learning for end-to-end autonomous driving,'' \emph{arXiv preprint arXiv:2403.02877}, 2024.

\bibitem{almalioglu2022deep}
Y.~Almalioglu, M.~Turan, N.~Trigoni, and A.~Markham, ``Deep learning-based robust positioning for all-weather autonomous driving,'' \emph{Nature Machine Intelligence}, pp. 749--760, 2022.

\bibitem{velas2018cnn}
M.~Velas, M.~Spanel, M.~Hradis, and A.~Herout, ``Cnn for imu assisted odometry estimation using velodyne lidar,'' in \emph{IEEE Int. Conf. Autonom. Rob. Syst. Compet.}, 2018, pp. 71--77.

\bibitem{wang2019deeppco}
W.~Wang, M.~R.~U. Saputra, P.~Zhao, P.~Gusmao, B.~Yang, C.~Chen, A.~Markham, and N.~Trigoni, ``Deeppco: End-to-end point cloud odometry through deep parallel neural network,'' in \emph{IEEE/RSJ International Conference on Intelligent Robots and Systems}, 2019, pp. 3248--3254.

\bibitem{li2020dmlo}
Z.~Li and N.~Wang, ``Dmlo: Deep matching lidar odometry,'' in \emph{IEEE/RSJ International Conference on Intelligent Robots and Systems}, 2020, pp. 6010--6017.

\bibitem{zheng2020lodonet}
C.~Zheng, Y.~Lyu, M.~Li, and Z.~Zhang, ``Lodonet: A deep neural network with 2d keypoint matching for 3d lidar odometry estimation,'' in \emph{ACM Int. Conf. Multimedia}, 2020, pp. 2391--2399.

\bibitem{liu2022lidar}
T.~Liu, Y.~Wang, X.~Niu, L.~Chang, T.~Zhang, and J.~Liu, ``Lidar odometry by deep learning-based feature points with two-step pose estimation,'' \emph{Remote Sensing}, p. 2764, 2022.

\bibitem{li2019net}
Q.~Li, S.~Chen, C.~Wang, X.~Li, C.~Wen, M.~Cheng, and J.~Li, ``Lo-net: Deep real-time lidar odometry,'' in \emph{IEEE Conf. Comput. Vis. Pattern Recog.}, 2019, pp. 8473--8482.

\bibitem{wang2021pwclo}
G.~Wang, X.~Wu, Z.~Liu, and H.~Wang, ``Pwclo-net: Deep lidar odometry in 3d point clouds using hierarchical embedding mask optimization,'' in \emph{IEEE Conf. Comput. Vis. Pattern Recog.}, 2021, pp. 15\,910--15\,919.

\bibitem{ali2023delo}
S.~A. Ali, D.~Aouada, G.~Reis, and D.~Stricker, ``Delo: deep evidential lidar odometry using partial optimal transport,'' in \emph{Int. Conf. Comput. Vis.}, 2023, pp. 4517--4526.

\bibitem{wang2022efficient}
G.~Wang, X.~Wu, S.~Jiang, Z.~Liu, and H.~Wang, ``Efficient 3d deep lidar odometry,'' \emph{IEEE Trans. Pattern Anal. Mach. Intell.}, 2022.

\bibitem{cho2020unsupervised}
Y.~Cho, G.~Kim, and A.~Kim, ``Unsupervised geometry-aware deep lidar odometry,'' in \emph{IEEE Int. Conf. Rob. Auto.}, 2020, pp. 2145--2152.

\bibitem{10906337}
G.~Wang, Z.~Feng, C.~Jiang, J.~Liu, and H.~Wang, ``Unsupervised learning of 3d scene flow with lidar odometry assistance,'' \emph{IEEE Trans. Intell. Transport. Syst.}, pp. 4557--4567, 2025.

\bibitem{cho2019deeplo}
Y.~Cho, G.~Kim, and A.~Kim, ``Deeplo: Geometry-aware deep lidar odometry,'' \emph{arXiv preprint arXiv:1902.10562}, 2019.

\bibitem{SelfVoxeLO}
Y.~Xu, Z.~Huang, K.-Y. Lin, X.~Zhu, J.~Shi, H.~Bao, G.~Zhang, and H.~Li, ``Selfvoxelo: Self-supervised lidar odometry with voxel-based deep neural networks,'' in \emph{Conference on Robot Learning}, 2021, pp. 115--125.

\bibitem{xu2022robust}
Y.~Xu, J.~Lin, J.~Shi, G.~Zhang, X.~Wang, and H.~Li, ``Robust self-supervised lidar odometry via representative structure discovery and 3d inherent error modeling,'' \emph{IEEE Rob. Auto. L.}, pp. 1651--1658, 2022.

\bibitem{low2004Linear}
K.~L. Low, ``Linear least-squares optimization for point-to-plane icp surface registration,'' \emph{Chapel Hill}, 2004.

\bibitem{nubert2021self}
J.~Nubert, S.~Khattak, and M.~Hutter, ``Self-supervised learning of lidar odometry for robotic applications,'' in \emph{IEEE Int. Conf. Rob. Auto.}, 2021, pp. 9601--9607.

\bibitem{li20233d}
C.~Li, F.~Yan, S.~Wang, and Y.~Zhuang, ``A 3d lidar odometry for ugvs using coarse-to-fine deep scene flow estimation,'' \emph{Transactions of the Institute of Measurement and Control}, pp. 274--286, 2023.

\bibitem{fu2022self}
X.~Fu, C.~Liu, C.~Zhang, Z.~Sun, Y.~Song, Q.~Xu, and X.~Yuan, ``Self-supervised learning of lidar odometry based on spherical projection,'' \emph{Int. J. Adv. Robotic Sys.}, vol.~19, no.~1, 2022.

\bibitem{zhou2023hpplo}
B.~Zhou, Y.~Tu, Z.~Jin, C.~Xu, and H.~Kong, ``Hpplo-net: Unsupervised lidar odometry using a hierarchical point-to-plane solver,'' \emph{IEEE Trans. Intel. Veh.}, pp. 2727--2739, 2023.

\bibitem{aloam}
T.~Qin, S.~Cao, J.~Behley, and C.~Stachniss, ``A-loam: Advanced implementation of loam,'' \url{https://github.com/HKUST-Aerial-Robotics/A-LOAM}, 2019, accessed: 2025-06-15.

\bibitem{ICP}
J.~Giseop~Kim, ``Pyicp slam.'' in \emph{\url{https://github.com/JustWon/PyICP-SLAM}}, 2023.

\bibitem{koide2021voxelized}
K.~Koide, M.~Yokozuka, S.~Oishi, and A.~Banno, ``Voxelized gicp for fast and accurate 3d point cloud registration,'' in \emph{IEEE Int. Conf. Rob. Auto.}, 2021, pp. 11\,054--11\,059.

\end{thebibliography}

\begin{IEEEbiography}[{\includegraphics[width=1in,height=1.25in,keepaspectratio]{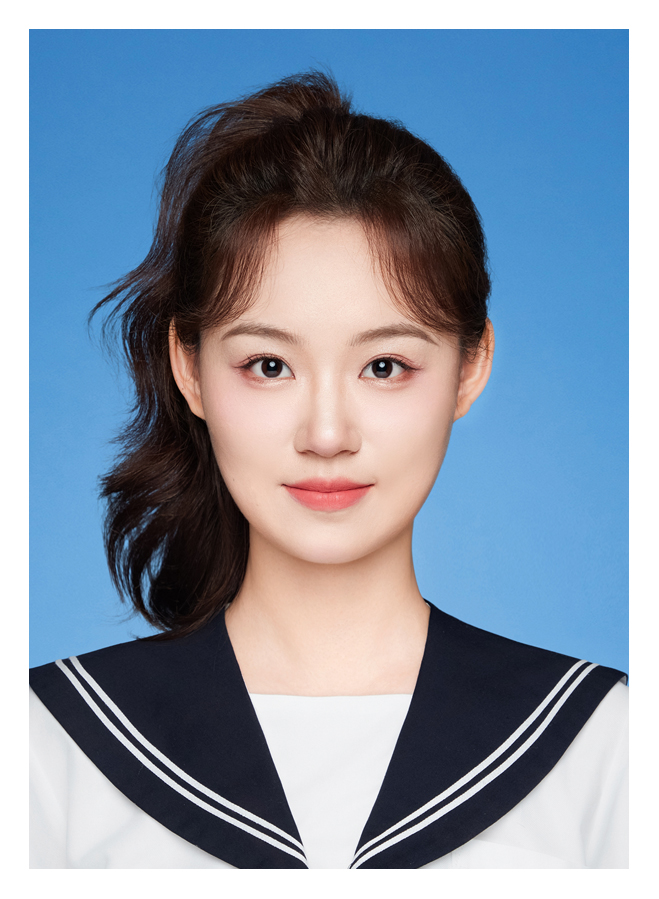}}]
 {Beibei Zhou} received her Ph.D. degree at the PCA Lab, he Key Lab of Intelligent Perception
and Systems for High-Dimensional Information of
Ministry of Education, School of Computer Science and Engineering, Nanjing University of Science and Technology, Nanjing, China, in 2024. She is currently working in Shanghai Polytechnic University. From 2023 to 2024, she was a visiting scholar with the State Key Laboratory of Internet of Things for Smart City (SKL-IOTSC), University of Macau, Macau.
 Her research interests include visual/Lidar odometry, 3D scene flow estimation, and computer vision.
\end{IEEEbiography} 
\vspace{-0.3cm}

\begin{IEEEbiography}[{\includegraphics[width=1in,height=1.25in,keepaspectratio]{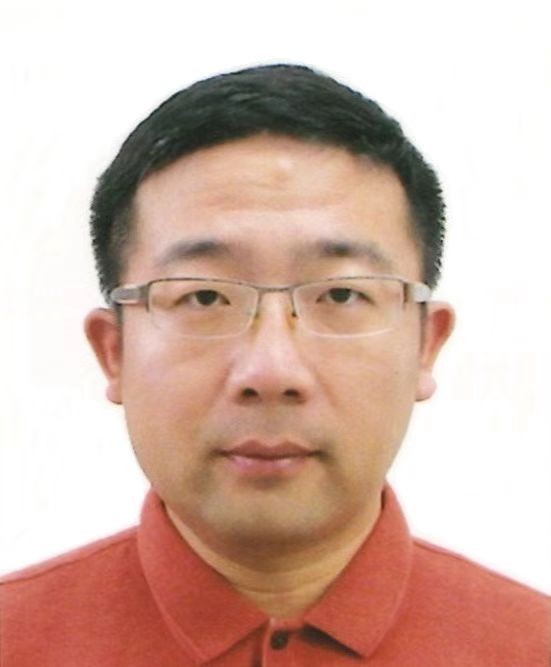}}]
 {Zhiyuan Zhang} received the Ph.D. degree in computer science from the National University of Singapore in 2015. He is currently an Assistant Professor at the School of Computing and Information Systems, Singapore Management University. Previously, he was an Assistant Professor at the Ningbo Innovation Center, Zhejiang University, and a Research Fellow at DMand Lab, Singapore University of Technology and Design. From 2014 to 2017, he worked as a Staff Researcher at Lenovo, leading computer vision projects. His research interests include 3D point cloud processing, lightweight deep learning, and 3D biometrics.
\end{IEEEbiography} 
\vspace{-0.3cm}

\begin{IEEEbiography}[{\includegraphics[width=1in,height=1.25in,keepaspectratio]{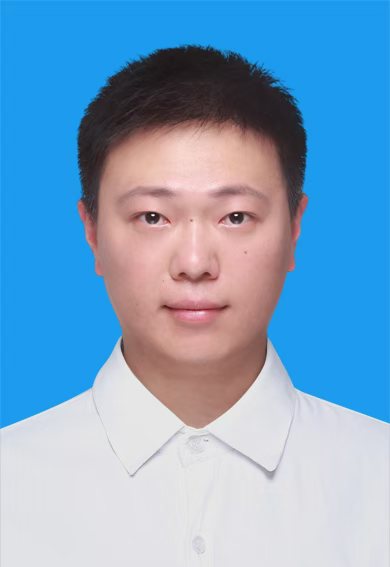}}]
 {Zhenbo Song} (Member, IEEE) received the Ph.D. degree in Control science and engineering from Nanjing University of Science and Technology, China, in 2022, where he is currently an Associate Professor. From 2018 to 2019, he was a Visiting Scholar with the Australian National University, Canberra, Australia. His current research interests include autonomous driving, image processing, 3-D reconstruction, and deep learning.
\end{IEEEbiography} 
\vspace{-0.3cm}

\begin{IEEEbiography}[{\includegraphics[width=1in,height=1.25in,keepaspectratio]{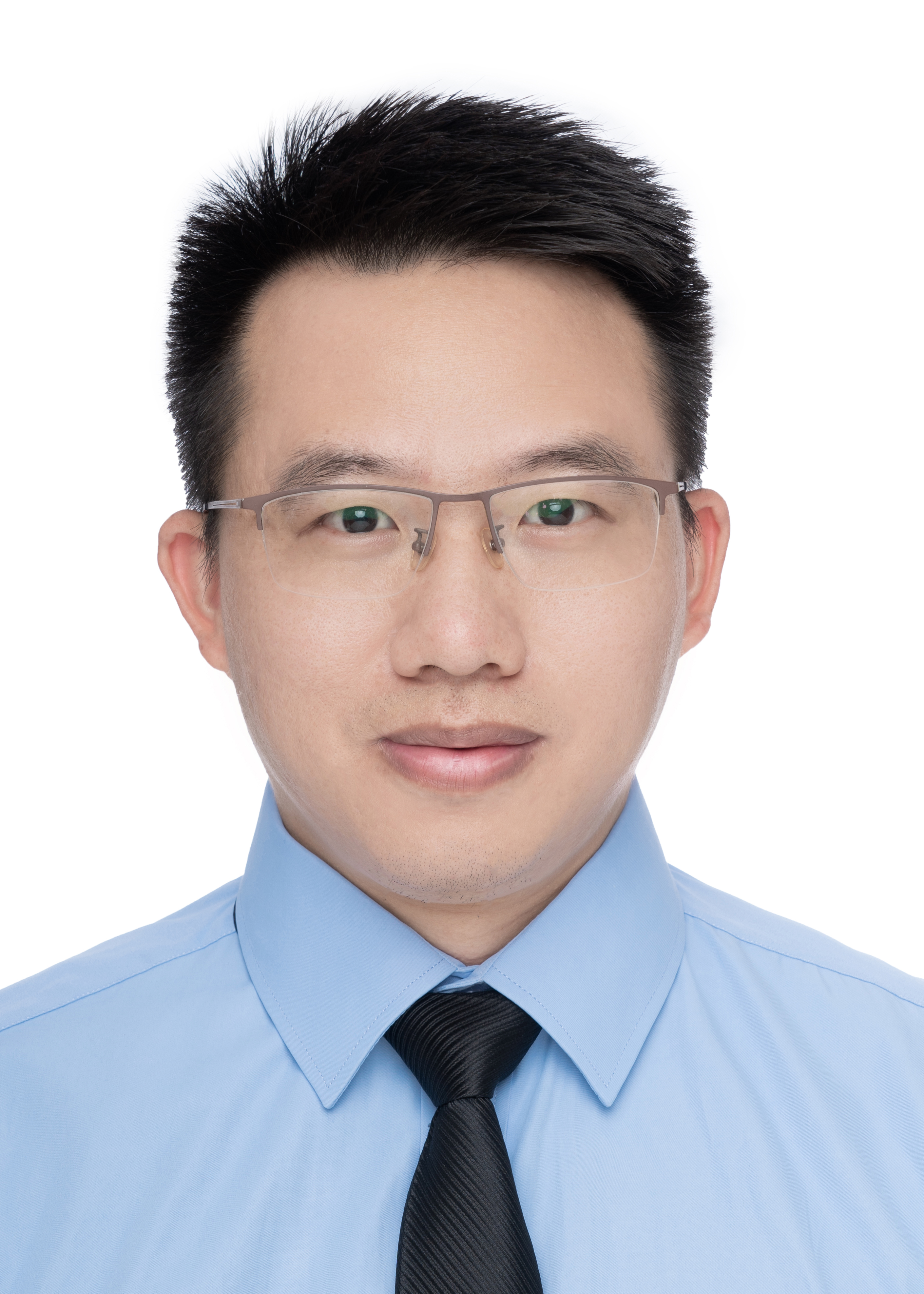}}]
 {Jianhui Guo} received the BS degree, MS degree and PHD degree from the Nanjing University of Science and Technology, Nanjing, China, in 2003, 2005 and 2008, respectively. In 2008, he joined the Nanjing Institute of Electronics Technology as a senior engineer.  He is now an associate professor in the School of Computer Science and engineering of Nanjing University of Science and Technology. His research interests include machine learning, data mining, pattern recognition, and intelligent robot, and information fusion.  (Email: guojianhui@njust.edu.cn)
\end{IEEEbiography} 
\vspace{-0.3cm}

 \begin{IEEEbiography}[{\includegraphics[width=1in,height=1.25in,keepaspectratio]{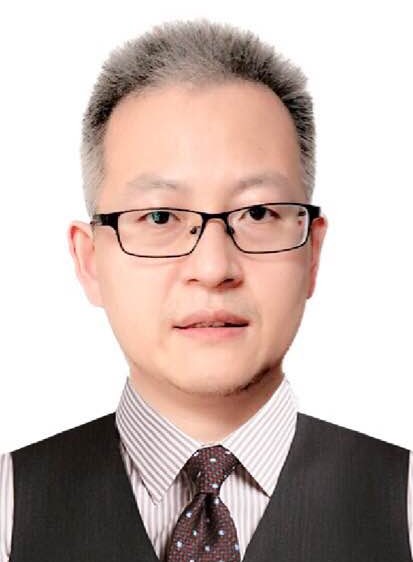}}]
 {Hui Kong}
received the Ph.D. degree in computer vision from Nanyang Technological University, Singapore, in 2007. He is currently an Associate Professor with the State Key Laboratory of Internet of Things for Smart City (SKL-IOTSC), Department of Electromechanical Engineering (EME), University of Macau, Macau. His research interests include sensing and perception for autonomous driving, SLAM, mobile robotics, multi-view geometry, and motion planning.
\end{IEEEbiography} 
\vspace{-0.3cm}

\bibliographystyle{IEEEtran}

\vfill

\end{document}